\documentclass[journal,onecolumn]{IEEEtran}

\usepackage{bm,amssymb,amsmath}
\usepackage{url}
\usepackage{caption}
\usepackage{subcaption}
\usepackage{color}

\newcommand{\lmanew}[1]{\textcolor{black}{#1}}

\usepackage[figuresright]{rotating}

\begin{document}

\title{Learning detectors of malicious web requests \\for intrusion detection in network traffic}

\author{
	\IEEEauthorblockN{\textbf{Lukas Machlica}$^\dag$, \textbf{Karel Bartos}$^{\dag,\S}$, \textbf{Michal Sofka\footnote{$^*$ This work was completed while M. Sofka was at Cisco Systems and at Czech Technical University, Faculty of Electrical Engineering, Czech Republic.}}$^{\ddag,\S,*}$
	\\
		\textit{lumachli@cisco.com, kbartos@cisco.com, msofka@4catalyzer.com} \\
	}
	\IEEEauthorblockA{
	$^\dag$Cisco Systems, Inc., Charles Square Center, Karlovo Namesti 10 Street, Prague, 12000, Czech Republic
	\\
	$^\S$ Czech Technical University in Prague, Faculty of Electrical Engineering, Czech Republic
	\\
	$^\ddag$ 4Catalyzer, 27 E 28th St, New York, NY 10016	
	}
}

\maketitle

\begin{abstract}
This paper proposes a generic classification system designed to detect security threats based on the behavior of malware samples. The system relies on statistical features computed from proxy log fields to train detectors using a database of malware samples. The behavior detectors serve as basic reusable building blocks of the multi-level detection architecture. The detectors identify malicious communication exploiting encrypted URL strings and domains generated by a Domain Generation Algorithm (DGA) which are frequently used in Command and Control (C\&C), phishing, and click fraud. Surprisingly, very precise detectors can be built given only a limited amount of information extracted from a single proxy log. This way, the computational requirements of the detectors are kept low which allows for deployment on a wide range of security devices and without depending on traffic context such as DNS logs, Whois records, webpage content, etc. Results on several weeks of live traffic from 100+ companies having 350k+ hosts show correct detection with a precision exceeding 95\% of malicious flows, 95\% of malicious URLs and 90\% of infected hosts. In addition, a comparison with a signature and rule-based solution shows that our system is able to detect significant amount of new threats.
\end{abstract}

\section{Introduction}
According to the recent threat intelligence statistics, malicious traffic is present on 100 percent of corporate networks~\cite{CiscoRep2014}. Growing proliferation of the attack model and the complexity of threats demand advanced solutions for detecting, classifying, and remediating new incidents. New and previously unseen polymorphic malware, zero-day attacks, or other types of advanced persistent threats are usually not detected or blocked by signature-based security devices, firewalls, or anti-viruses \cite{Bailey2007,ZhangBailey2013}. 

The network traffic can be classified at different levels of detail. The approaches based on packet inspection and signature matching rely on a database of known malware samples and manually designed expert rules \cite{ModiJNCA2012,LiaoJNCA2012}.
In order to keep the number of false alerts low, the rules have to be strict such that they do not match legitimate traffic. This limits the detection ability of these systems and requires frequent updates of the rules and the sample databases \cite{MaICML09}.
Moreover, due to the continuous improvements of network bandwidth, analyzing individual packets is becoming intractable on high-speed network connections. It is more efficient to classify network traffic based on flows representing groups of packets, e.g.\ NetFlow \cite{netFlow} or proxy logs \cite{Lou02cut2002}. The proxy logs record all communication over HTTP, a protocol becoming very popular by malware \cite{Sandnet,Perdisci2010}.

The methods based on statistical features extracted from the proxy log fields have shown the promise of detecting malware behaviors of different malware families \cite{Phoenix,DomFlux2012}. The detection algorithms rely on the fact that an adversary needs to communicate with the infected host.
For example, in phishing or click fraud, stolen credentials or sensitive private data are transfered to the botmaster. The botmaster might use a pull style Command and Control (C\&C) to download (pull) commands from remote servers by the bots. To protect the C\&C servers and avoid the detection, malware uses different detection evasion techniques such as domain-flux \cite{FluxNDSS08,DomFlux2012,GueridMS13} or domain generation algorithms \cite{Antonakakis2012,Phoenix}. 
Malware might also encrypt different parts of the URLs \cite{Garcia2014}, or use generated or encrypted strings in the URL path or the URL query to avoid the detection by signatures based on regular expressions~\cite{Rovnix}. These techniques complicate the identification of the botmaster, but the malware communication patterns can be still used to discover the infected hosts.

This paper presents a system for detecting malware behaviors by reusing modular system components applicable to a wide range of malware. This is achieved by a generic learning-based architecture that is trained to detect network communication of various malware families using a database of malware samples. The learning relies on statistical features computed from the proxy log fields. 
The system monitors the traffic for malicious communication with infected hosts which makes it possible to detect various malware behaviors.
The infected hosts are identified using features from a single proxy log. Although this can be limiting when discovering attacks over a period of time (e.g.\ DDoS), it is less restrictive than clustering-based approaches that require multiple hosts to be infected with the same malware \cite{BotSniffer,Phoenix}.
The features extracted on a dataset of training malware communication samples are used in a random forest learning algorithm to train detectors for a set of predetermined behaviors. 
Since various malware types can share specific behaviors, the detectors serve as basic building components of a multiple level detector hierarchy. Each path through the detectors hierarchy finds a specific malware manifestation while reducing false positives, increasing robustness, and improving computational speed.

Our detectors are trained to find malware communication patterns from encrypted URL strings and their specific distribution of characters, which are most common for malicious activities related to C\&C, phishing, and click fraud.
One of our detectors is also trained to find malware communication using automatically-generated malicious domains. As will be shown, these domains can be detected with high precision even without harvesting additional intelligence from the DNS and Whois records, the web page content or behavior correlation (see Section~\ref{sec:rel-work} with related works).
This approach can be seen as a data-driven alternative or complement to signatures for cases, where signatures are hard to define. Moreover, since we use statistical features that are invariant to behavior changes up to a certain degree, the ability of the detectors to generalize to new and previously-unseen malware is higher in comparison to signatures. The precision and recall of our system can be further increased by an integration with intrusion detection systems that use contextual information \cite{BotMiner,KurtISSP2011}.

In summary, this paper presents the following major contributions.
\begin{itemize}
\item Scalable multi-level learning architecture uses generic features and can be adopted for \textit{precise detection of different types of malware behavior} (e.g.\ C\&C, phishing, click-fraud) without designing malware-specific rules or heuristics.
\item The detectors only process \textit{individual proxy logs without any additional information} related to the malware executables, DNS records, or HTML content, which decreases the cost of collecting additional intelligence and increases the detection speed. The detectors do not store any traffic-related states, time-dependent statistics, or trend observations and therefore the detected threats can be instantly reported and mitigated.
\item The hierarchy \textit{increases the computational speed and robustness} since the earlier detection stages use simpler, easier-to-compute features to classify legitimate behaviors. The later stages can then focus on classifying harder cases but already in a search feature space reduced by the previous stages.
\item The modularity of the architecture allows for \textit{streamlined updates when new malware samples or behaviors are discovered} simply by (re)training the classifiers in different levels of the detection hierarchy.
\item \textit{Experiments are performed on live traffic} collected over one year, spanning 100+ companies, 350k+ hosts and more than 200 billion flows. In a 6 week evaluation period, the system was able to detect over 200 thousand malicious flows to 5000 different infected domains related to the infection of more than 4000 users. The system runs in real-time and is scalable by including additional computation nodes.
\end{itemize}

\section{Motivating Examples and Malware Types}
\label{sec:motivation}
This section gives examples of behaviors that are automatically recognized by the proposed system. These behaviors are commonly found in modern malware and therefore form the focus of our analysis in this paper. However, the system is general and can be easily extended to recognize additional behaviors, see Section~\ref{sec:mal-class}.

Various types of malware behavior can be distinguished by observing their communication patterns with the command and control server. The patterns for the same malware behavior will use proxy logs with similar attributes which can be exploited to detect particular behavior. For example, malware can use domains generated automatically by an algorithm, traffic timings, or data encryption. Since malware can have a combination of these properties and behaviors, the strategy taken in this paper is to model each of them separately. This increases the generalization capability of the malware classifier since it focuses on exhibited known behaviors and their combination rather than targeting specific malware types.

\subsection{Encrypted Strings in URLs}

Malware may establish a command-and-control channel, track user activity, have rootkit capability, and perform click-fraud through the automatic loading and clicking of unsolicited advertisements. The attacker may obtain information about the infected device and attempt to further exploit the device with additional threats. Threat communication over HTTP with its command-and-control infrastructure may use a domain generated by a Domain Generation Algorithm (DGA). The communication may also employ encryption algorithms to obfuscate the transmitted information~\cite{Garcia2014}. The information transfered as a part of the URL can contain malware instructions to be executed on the infected host such as pieces of JavaScript, malware parameters, or threat configurations. 

Existing malware detection techniques, searching for information transfered in the URL string, rely on rules and regular expressions to capture specific URL patterns or tokens \cite{urlpatterns,MaACM2011}, see Section~\ref{sec:rel-work} for examples. These techniques become ineffective when the pattern changes as a result of malware update or obfuscation (the Rovnix malware uses generated URL tokens to avoid pattern matching~\cite{Rovnix}). Detectability of these threats is further complicated by legitimate HTTP requests that contain encrypted strings in a part of the URL. For example, CSS image sprites, Base64 encoding of strings that contain URI reserved or unsafe characters \cite{URI}. Few examples of malicious and legitimate URLs with encrypted strings are given in Table~\ref{tab:tunn-ex}.

\begin{table}
	\centering
	\caption{
		Example of malicious and legitimate URLs with encoded strings. The malicious URLs are related to Dridex, Palevo worm and Alureon Trojan.
	}
	\begin{tabular}{l}
		\hline
		\textbf{malicious} \\
		\hline
		\textit{hxxp:}//lkckclckl1i1i[.]com/zKG2ZB1X6M5xt5c1Y2xrPTEuNyZiaWQ9..
		\\
		\textit{hxxp:}//109.235.54[.]162/m/IbQEZVVjipFdkB0KHeNkNuGBabgSr2z3..
		\\
		\textit{hxxp:}//78.140.164[.]160/jjlqXpP/$\sim$GIja7A3q/KqRSx+1s8kNC=/\%2BsI..
		\\
		\textit{hxxp:}//masterproweb[.]net/images2/BD3006FB490CADF111E40696D3..
		\\ \hline
		\textbf{legitimate} \\
		\hline
		\textit{http:}//www.thespec.com/DependencyHandler.axd/L0Rlc2t0b3BNb2R1b..
		\\
		\textit{http:}//www.degruyter.com/assets/virtual/H4sIAAAAAAAAAKWSwUo..
		\\
		\textit{http:}//www.1001pneus.fr/ls/YToxNzp7czo0OiJwYWdlIjtzOjY6InNlYX..
		\\ \hline
	\end{tabular}
	\label{tab:tunn-ex}
\end{table}

\paragraph{Phishing And Click-Fraud}
A special case of malware with encrypted strings as a part of URL are phishing and click-fraud, where in some cases encrypted strings are used to hide the transfer of client credentials or redirection instructions. Even if the initial URL shown to the user looks benign, some of the requests of the communication that follows can expose the attacker. Specific examples that we aim for are given in Section~\ref{sec:phish-exp} and Section~\ref{sec:clickfraud}.

\subsection{Domain Generation Algorithm (DGA)}
\label{sec:dga-ex}
Using domain black lists to block malware communication is complicated when the botnets use domains generated by a Domain Generation Algorithm (DGA). DGAs dynamically create a large list of domains out of which only a small subset is actually registered. The same DGA is used on all infected machines with a random seed derived from the system date. The botmaster uses the knowledge of the DGA to register a predetermined subset of the generated domains which are then used to establish a C\&C channel. DGAs are configurable and the botmaster can manipulate the way the domains are generated. This makes the use of detection mechanisms relying on static black lists infective since it is not possible to keep the lists up-to-date with the large amounts of automatically generated domains \cite{Antonakakis2012}. As an alternative, domain lists can be easily compiled when the DGA is known. However, this requires reverse engineering of the malware sample which is not always feasible \cite{Antonakakis2012}.

Evolving and configurable DGAs are becoming sophisticated to the point that they can bypass existing detection algorithms. Approaches based on the rules derived from the distribution of characters in the domain names are frequently ineffective and may produce many false detections. One example of falsely detecting domains as generated by a DGA are domains including government institutions, which are often initialisms of the institution names. Another example are domains with some vowels left out, or proxy and firewall checks of legitimate applications (e.g.\ web browsers), which are implemented as HTTP requests to non-existing generated domains. Additional intelligence derived from the DNS and WHOIS records is sometimes used to reduce the number of false detections \cite{Antonakakis2012,Exposure}.
Example of malicious domains from DGA and legitimate domains, which can be possible source of false detection, are given Table~\ref{tab:dga-ex}.

\begin{table}
\centering
\caption{Example of malicious generated domains related to Gamarue/Andromeda botnet, Geodo botnet and Emotet banking Trojan and legitimate domains. The last legitimate domain is a village name in NW Wales.
}
\begin{tabular}{c}
\hline
\textbf{malicious} \\
\hline
hzmksreiuojy[.]in, b9qmjjys3z[.]com, jaoohqvqda[.]ru, \\
oqjiwef12egre6erg6qwefg312qrgqretg132[.]com, lkckclckl1i1i[.]com, \\
xjpakmdcfuqe[.]nl, reqblcsh[.]net, cilavocofer[.]eu
\\ \hline
\textbf{legitimate} \\
\hline
skhhtcss.edu.hk, edkowalczyk.com, blkdmnds.com, \\
watdoejijbijbrand.nl, kdnlrklb.com, abcdefgtfddf2223.com, \\
llanfairpwllgwyngyllgogerychwyrndrobwll-llantysiliogogogoch.com
\\ \hline
\end{tabular}
\label{tab:dga-ex}
\end{table}

\section{Challenges}
\label{sec:requirements}
Our data-driven malware classification approach has to overcome several challenges to make it practical in today's enterprise security environments.

\subsection{Precision}
\label{sec:acc}
Techniques based on signatures and regular expressions can be effective in providing low number of false positives, but they do not have a generalization capability \cite{ModiJNCA2012}. For example, once the URL in the proxy log changes (as a result of malware being updated), the regular expression can no longer capture this malicious communication.

Classifiers that are automatically trained using statistical features extracted from proxy logs have much higher generalization power. This is because the training procedure automatically finds rules and features that are characteristic for a particular malware behavior considering all the provided legitimate and malicious samples. The challenge is to keep the false positives low since some legitimate traffic proxy logs might appear similar to the malicious logs, see previous section.

When designing the system, the precision of the detector is typically evaluated on the level of URLs or proxy logs, i.e.\ using number of correctly and incorrectly classified URLs or logs. Although the incident report might contain the malicious and suspicious logs, the main interest to the security analysts are the infected hosts or user machines. It is therefore important to also evaluate the precision at the level of hosts. Otherwise, it can happen that a very accurate detector at the proxy log level, correctly classifying hundreds of thousands of proxy logs as malicious, can incorrectly report many hosts as infected. The discrepancy would be caused by many malicious flows originating at a few hosts but a handful of false positive flows distributed across many users. This issue will be further discussed in Section~\ref{sec:eval}.

\subsection{Adaptability}
\label{sec:adapt}
The malware detection system has to achieve high precision and recall and retain it after deployment. Therefore, there needs to be a mechanism to update the detectors in a reaction to emerging malware patterns. In data-driven systems, the update can be performed by automatically retraining the detectors using new malware traffic samples. It is clear that maintaining a larger set of classifiers updated at all times is hard. One way to address this problem is to design system components that focus on specific manifestations of malware behaviors. The components can then be reused when classifying behaviors from different malware families.

The system should also be able to update the detectors in the event that some of their parameters are compromised. For example, an adversary might discover particular domain or URL patterns that are not detected as malicious but can be used to communicate with a C\&C server. If this happens, the system needs to update its set of parameters, change input-output relationships, or update configurations, such that the accuracy is preserved but the evasion technique is not possible anymore. 
Our system is updated by randomized training that changes the set of learned parameters in each training run, see Section~\ref{sec:rf}. Retraining the classifier will result in a different set of parameters with focus on different features.

\subsection{Scalability}
\label{sec:scala}
The ever rising variability of malware increases the complexity of the detectors and their running time~\cite{LiaoJNCA2012}. The proxy log analytics system must be able to process the incoming data stream in real time to generate up-to-date reports and fulfill 24/7 availability in the business level agreements.

The number of available malicious samples cannot be a limiting factor, when creating or updating the detectors \cite{Zhang2013}. All available intelligence should be used when developing a detection algorithm. The update of the system to handle new malware behaviors needs to be seamless. The deployment of the update and validation of the reported incidents must be streamlined to keep the maintenance costs down. The system must be able to scale to large number of customers with tens of thousands or hundreds of thousands users~\cite{zhangTIFS2014}.

\subsection{Data-Related Challenges}
\label{sec:restr}
These restrictions are related to individual aspects of the problem definition. 

\paragraph{Proxy logs} Since we work with proxy logs, which consist of fields represented by numerical values (e.g. transfered bytes), strings (e.g. URL), categorical (http status) and other, the system has to support different types of features (binary, continuous, categorical, ordinal assessment, etc.) and address the problem with different scalings of each feature. It has to deal with missing values during the runtime since various fields from proxy logs can be missing.

\paragraph{Data sources} The system should be able to make use of incomplete training datasets. It primarily works at the proxy log level, but since no extensive datasets with labeled threats on the proxy log level are available, it should be able to make use of datasets containing only partial information: e.g. domain-names, URLs, regexp patterns related to a specific threat. Once a new malware pattern is discovered, it should be possible to incorporate new features related to the pattern. This should not result to an expensive reconstruction of the whole system. 

\

In order to deal with the challenges, we use a set of generic features (Section~\ref{sec:fextr}). The features are used to represent string patterns by the distribution of characters in the URL and thus generalize better to new malware than signature-based solutions (Section~\ref{sec:mal-class}). These features are further augmented by additional proxy log fields such as the number of transfered bytes, MIME-type and HTTP status. We use random forest discriminative classifier (Section~\ref{sec:rf}), which can effectively deal with different features types and provides estimates of their importance. The random forest can be easily (re)trained, the detection costs are low (Section~\ref{sec:costs}), and it is robust against missing values. The multi-level architecture, introduced in Section~\ref{sec:arch}, can be thought of as a chain of base classifiers, where each base classifier is represented by a random forest. The architecture is used to gradually focus the classification on fewer flows and differentiate between different malware behavior types such as C\&C communication, phishing or click-fraud. The maintenance of the system is lowered by sharing the detectors at particular levels (see Fig.~\ref{fig:sys-full}).
The architecture can also be used to identify potentially malicious samples that share a similar communication pattern, e.g.\ encrypted strings in the URL path. The samples would then be analyzed and included in training of new detectors (see experiments conducted at the end of Section~\ref{sec:ENC}).

\section{Architecture}
\label{sec:arch}
We use a simple, yet powerful and scalable classification architecture that effectively deals with the issues discussed in Section~\ref{sec:requirements}. Instead of training a single complex classifier, we train several base classifiers. They are trained using various sets of features and datasets in order to represent different aspects of the communication. The final classifier is constructed as a multi-level arrangement of the base classifiers. Note that in cases where the difference is not obvious, we use the term \emph{base classifier} to denote individual levels of the multi-level classifier. Let us now elaborate on the capabilities and advantages of the architecture.

\subsection{Base Classifiers}
The classifiers in our architecture are constructed as a multi-level arrangement of base classifiers.
Base classifier is a binary classifier that represents one building block in the architecture. Formally, it is a function that takes a sample $x$ as input and outputs the conditional probability $p(b | x)$ that the sample $x$ is part of the behavior $b$ represented by this function/classifier (e.g. communication with encrypted strings in the URL). We say that the sample $x$ was detected by a classifier $c_b$ responsible for the behavior $b$ if $p(b | x) > \tau$, where $\tau$ is the detection threshold. 

Each of the base classifiers can be trained independently, or it can be related to an output of another classifier. In the latter case we use two different ways how to build the classifiers:
\begin{itemize}
\item Bottom-up. The base classifier at a certain level complements the decision of a base classifier at the previous level. The classifier on a higher level is built only from samples that were detected at the previous level. That is, the amount of detected samples decreases at each level and the precision gets increased. Each base classifier is trained from a specific sample and feature set that capture important characteristics of the traffic we would like to detect at a certain level. This narrows the scope of the classifier and is likely the most important step towards fewer mistakes~\cite{Sommer2010}. 
\item Top-down. A high-precision classifier is first built on the last level. We then make use of the constraint relaxation technique to remove some samples and/or features from the training set. This constructs the preceding classification level with the goal to increase recall.
\end{itemize}
\begin{figure}
\centering
\includegraphics[width=0.4\linewidth]{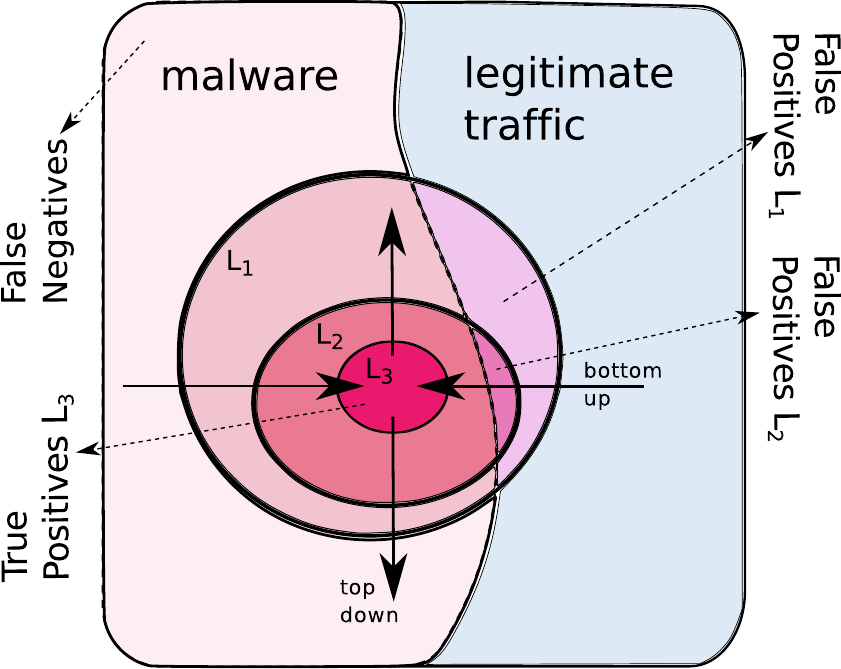}
\caption{Illustrative example of the building process and multi-level detection. Each additional detection level ($\text{L}_1\rightarrow\text{L}_2\rightarrow\text{L}_3$) covers a subset of behavior from previous level. Higher levels focus on more specific malware behaviors. That is, precision increases at each level, often at the cost of recall. The multi-level system can be constructed from the highest level L$_3$ relaxing the constraints, increasing recall and adding lower levels (top-down). Or it can be build from bottom describing general behavior toward exact malware manifestations (bottom-up).}
\label{fig:build}
\end{figure}
Example that illustrates both approaches is depicted in Figure~\ref{fig:build}. Note that the multi-level design also facilitates the interpretation of the classifier's output by gradually labeling and narrowing the type of the traffic at each level.

The bottom-up training is especially useful when the number of labeled training samples is low, but they are connected through a simple pattern for which enough data can be cheaply collected and therefore can be easily learned. Let us demonstrate the idea on an example. When a DGA classifier is trained, it is relatively easy to collect malicious domains from DGA (e.g. from unsuccessful DNS requests, reverse engineering, etc.), but it is much more expensive to collect full proxy logs of malicious DGA requests. Classifier based on the domain names only, does not yield sufficient precision, cannot be used alone, but serves as a good prefiltering tool discarding obvious non DGA domains. It will create boundaries in the domain name space, extensively reduce the number of samples needed to be processed by the next level, and thereby reduce the number of samples needed to train the classifier at the subsequent level. Moreover, reduced output of the first level can be much more easily inspected for new malicious DGA samples than the full traffic.

The top-down approach can be used when a working prototype of a (multi-level) classifier was build, but the time consumption related for example to the extraction of features at the first level is high. Once such an issue is discovered, rather than removing features that are costly or replacing the classifier with a cheaper one, a simpler base classifier (in terms of calculation costs - only a few cheap features are used) is trained and placed on the first level in the classification chain. We can control the trade off between recall and time consumption of the multi-level classifier by setting up the working point (in the sense of recall and precision) of the first base-classifier. With strict first level we will get only a few samples processed at the other more expensive levels. This will result in decreased time consumptions, but also lower recall, the opposite is true for a vague first-level classifier. The operating point can change in time according to the actual processing load.

\subsection{Advantages}
Let us mention several other important properties of the multi-level architecture:

\begin{itemize}
\item Since the number of detected flows is dramatically reduced at each level (see Section~\ref{sec:mal-class}), the comparison of inputs and outputs of each level requires less effort than analyzing the input data. This makes it easier to find new malware samples and to train new classification levels or retrain existing levels (see experiments performed in Section~\ref{sec:ENC}).
\item Each base classifier can be build from a subset of samples and features. This is used to effectively solve the problem with incomplete training datasets stated in Section~\ref{sec:restr}. For example, a URL can be regarded as a proxy log with missing values. Having a set of proxy logs and a set of URLs, we can easily utilize information from both building different classification levels for each source separately (see the training of individual levels discussed in Section~\ref{sec:dga-tr}).
\item Building blocks can be reused through sharing of classification levels between different classifiers, see~Fig.~\ref{fig:sys-full}. Once new malware samples are available or a new significant malware pattern is discovered, responsible detector is retrained or new detector is added at a specific level. If such level is shared across different malware classifiers, the update of a particular malware classifier contributes to several classifications at once.
\end{itemize}

\section{Discriminative classifier}
\label{sec:classifier}
We have chosen the random forest classifier \cite{Breiman2001RF, Criminisi2011TR} as the implementation for the base classifier. It is a discriminative classifier based on randomized binary decision trees. \lmanew{It has several desired properties related to our problem:}
\begin{itemize}
\item \lmanew{it can process features of multiple types (e.g. real-valued, binary, categorical; see Section~\ref{sec:fextr}) without additional data preprocessing - e.g. the data do not have to be normalized, this is often crucial for other standard algorithms from machine learning} \cite{Bishop2006} 
\item it is easy to train and runs efficiently on large datasets
\item it maintains accuracy even when large proportion of data is missing, i.e. problem of missing values
\item can learn non-linear models and model complex behavior
\item it provides estimates of importance of variables used in the classification, i.e. once features were designed, during the training of random forest non-informative features are discarded and the discriminative are promoted
\end{itemize}
\subsection{Binary Decision Tree}
\begin{figure}
\centering
\includegraphics[width=0.4\linewidth]{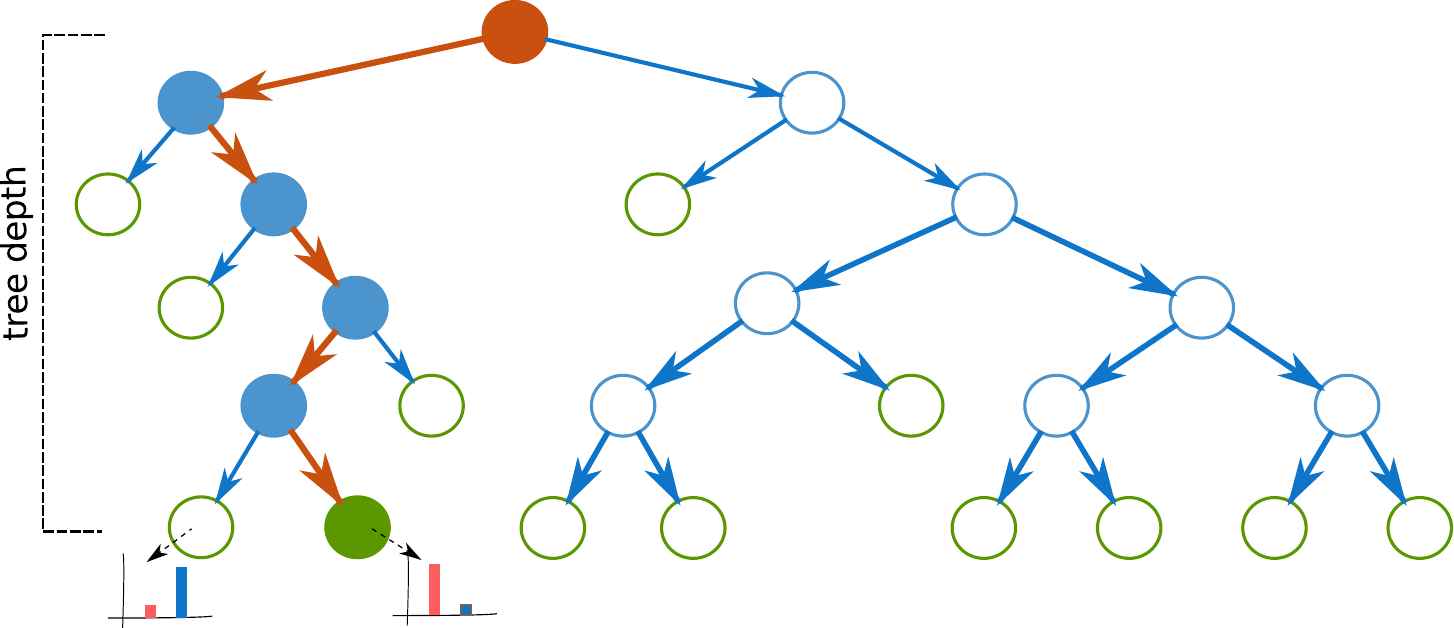}
\caption{Binary Decision Tree. Each split node (blue node) is linked with two other child nodes. The tree is traversed from the root (brown node) according to simple thresholding until the leaf (green node) is found. The classification is given by the class with majority of samples. This is depicted by histograms of class labels (two classes: blue vs red) below the leafs.}
\label{fig:tree}
\end{figure}

Let $\bm{x} = [x_1,\ldots, x_K]^\top$ be a $K$ dimensional feature vector. A binary decision tree is a collection of nodes $n_i$ organized in a hierarchical tree structure. Node can be a split node or a terminal leaf node, where to each split node exactly two child nodes are connected. To each split node $n_i$, feature index $\pi_i \in \{1, \ldots, K\}$ and splitting function $f(\bm{x}, \pi_i, \phi_i): \mathcal{R} \rightarrow \{0,1\}$ is associated:
\begin{equation}
f(\bm{x}, \pi_i, \phi_i) = 
\left\{
\begin{array}{l}
1 \text{ if } x_{\pi_i} > \phi_i
\\
0 \text{ if } x_{\pi_i} \leq \phi_i
\end{array}
\right.,
\label{eq:rf-f}
\end{equation}
i.e. $\pi_i$-th dimension of $\bm{x}$ is compared to a threshold $\phi_i$ in the node $i$. Based on the output of the splitting function the left, for $f(\bm{x}, \pi_i, \phi_i)=1$, or right branch, for $f(\bm{x}, \pi_i, \phi_i)=0$, of the tree is traversed. An example of a tree is given in Fig.~\ref{fig:tree}. 

\paragraph{Training}
The tree is built from the root. Given a sample set $X_i = \{\bm{x}_1, \ldots, \bm{x}_{N_i}\}$ of $N_i$ samples and set of adjacent labels $Y_i = \{\bm{y}_1, \ldots, \bm{y}_{N_i}\}$ available at node $n_i$, the threshold $\phi_i^*$ and feature index $\pi_i^*$ for node $n_i$ are chosen according to:
\begin{equation}
\phi_i^*, \pi_i^* = \arg \underset{\phi, \pi}{\text{max}} \: I(X_i, \phi, \pi).
\label{eq:rf-crit}
\end{equation}
The information gain $I(X_i, \phi, \pi)$ is given as
\begin{align}
I(\phi, \pi, X_i) &= H (X_i) - \sum_{b \in \{l, r\}}\frac{{N_{i,b}}}{{N_i}} H (X_{i,b}(\phi, \pi)), 
\notag
\\
X_{i,l}(\phi, \pi) &= \{\bm{x}_{j} \: | \: f(\bm{x}_j, \pi_i, \phi_i) = 1 \}, 
\notag
\\
X_{i,r}(\phi, \pi) &= \{\bm{x}_j \: | \: f(\bm{x}_j, \pi_i, \phi_i) = 0 \},
\notag
\end{align}
where the Shannon entropy $H(X)$ is computed on the label set $Y$ of $X$; $X_{i,l}(\phi)$, $X_{i,r}(\phi)$ are the sample sets after the split for left and right child node of $n_i$, and $N_{i,l}$, $N_{i,r}$ are their cardinalities, respectively. The information gain is maximized when the subset after the split contains samples from one class only, and it is minimized if the label distribution is uniform. 

The splitting ends when a predefined depth $D_\text{max}$ is reached or only labels of one class are present in the sample set of a (leaf) node.

\paragraph{Classification}
Once the classifier is trained, that is for each node $n_i$ optimal values $\pi_i^*$, $\phi_i^*$ were found, given a feature vector $\bm{x}$, the tree is traversed according to \eqref{eq:rf-f} until a leaf is found. We use the majority voting rule: label of the leaf is derived from the class with majority of training samples that finished in this leaf. If the count is same for all classes, the label is chosen randomly.

\subsection{Random Forest}
\label{sec:rf}
Random forest is a collection of $T$ randomized binary decision trees. Different parts of training process of a tree can be randomized \cite{RF-extreme, Mondriadn2014}. We use the following training strategy.
\begin{itemize}
\item Bagging. Given a training set $S_\text{train}$, for each tree the building set is sampled uniformly and with replacement from $S_\text{train}$. That is, the building set is expected to have approximately 63\% of unique samples from $S_\text{train}$. This technique is used to make individual trees more independent in the decision making.
\item Random feature selection. When the splitting criterion~\eqref{eq:rf-crit} is optimized at node $n_i$, we select $F$ feature indexes (dimensions of the feature vector) randomly, instead of taking all the $K$ features. Only features related to these indexes are used in the optimization for $n_i$. This makes the forest robust to missing data since the trees in the forest cannot rely on the full feature set. In addition potential overfitting is suppressed, since only partial information is utilized at each node \cite{Breiman2001RF}. 
\end{itemize}

To get the final classification score $s$, we sum up the decisions of each tree for each class and normalize by the number of trees. For example, in the two class classification problem with classes: malware with class label 1 and 0 for legitimate behavior, the overall score $s(\bm{x})$ in the forest for a sample $\bm{x}$ is determined according to:
\begin{equation}
s(\bm{x}) = \frac{1}{T} \sum_{t=1}^T d_t(\bm{x}),
\label{eq:rf-sc}
\end{equation}
where $d_t(\bm{x}) \in \{0,1\}$ is the decision of the $t$-th tree. The sample is malicious if $s(\bm{x}) > \tau$, where $\tau$ is a given (detection) threshold. The number of trees influences the precision of the decision. If the trees are independent so are the false positives, which are averaged out in \eqref{eq:rf-sc}.

\paragraph{Hyperparameters}
\label{sec:hyppars}
Hyperparameters of the random forest are: number of trees $T$, maximal depth of a tree $D_\text{max}$, and number of features $F$ randomly selected at a split node. In order to determine these hyperparameters, mostly cross-validation is utilized \cite{Criminisi2011TR}.

\section{Feature Extraction}
\label{sec:fextr}
\begin{figure}[t!]
\centering
\includegraphics[width=0.8\linewidth]{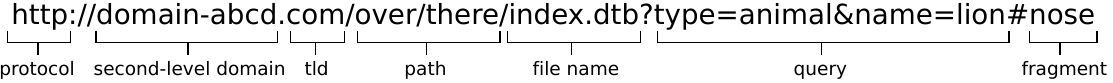}
\caption{Decomposition of URL into logical parts.}
\label{fig:url-decomposition}
\end{figure}

The system extracts one feature vector for each proxy log. \lmanew{The features are used to capture various malware behaviors. They are related to the statistical properties of characters present in the URL and statistical properties of the other proxy log fields. The goal is to define a large set of different features and to rely on the feature selection property of random forests to choose the most relevant features during training.}
More specifically, each URL is decomposed into seven logical components: protocol, second-level domain, top-level domain, path, file name, query, and fragment, see Figure~\ref{fig:url-decomposition}. (Since the transfered request body is not analyzed, the query strings in the proxy log URL are available only from HTTP GET requests.) A set of functions defined in Section~\ref{sec:feat-url} is then applied on components: domain, second-level domain, top-level domain, path, query and filename. Besides that, the same set of functions is also applied to the HTTP referrer. \lmanew{We also exploit ideas from the field of natural language processing and use  features based on the probability distribution of 3-grams (sequence of three characters) for the representation of strings of URL parts. Specifically, we use this representation for the second level domain strings when detecting domains generated by a DGA. We also use this representation for path and query strings when detecting encrypted strings in the URL.}
Features are extracted also for the other proxy log fields: flow duration, number of bytes transferred from client to server and from server to client, server and client port, user agent, MIME-Type, and HTTP status. These are defined in Section~\ref{sec:feat-flow}, and serve as additional support when distinguishing between malware and legitimate communication.

\subsection{Features for URL components}
\label{sec:feat-url}
Given a string $s$, we compute 26 features that characterize the use of specific patterns in the input string:
\paragraph{Length (1 feature)} $l(s)$ = number of characters in the string.
\paragraph{Ratio of consonant to vowel change (1 feature)}
$r_{c \rightarrow w}(s) = \frac{number\,\,of\,\,changes\,\,from\,\,consonant\,\,to\,\,vowel}{l(s)}.$  \\ The frequency of the changes between consonants and vowels is useful to differentiate ordinary words formed from syllables of strings representing abbreviations, encodings or generated names.
\paragraph{Maximum occurrence-ratio of a character (1 feature)} For each character in the string $s$, the number of occurrences of this character is divided by $l(s)$, and the maximum of these values is used. High frequency of a character in a string, e.g.\ in the query of a URL, is not common and can be seen as anomalous and discriminative.
\paragraph{Maximum occurrence-ratio for a character type (3 features)} We distinguish the following character types: number, an upper case and a lower case letter. For each of these types we compute the maximum occurrence-ratio in the same way as the maximum occurrence-ratio of character. These features represent the type of the string: whether it is mostly numerical, encrypted (encrypted strings often have many upper case characters) or in an ordinary common form (lower case).
\paragraph{Maximum length of stream of characters (3 features)} Computed as the length of a substring of $s$ containing only characters of the same type: lower case, upper case letters, digits. This feature complements the previous feature since it can differentiate between strings such as: "ver123456" and "r12v34e56". The previous feature would give the same occurrence-ratio of numbers for both strings, but it would not distinguish between these two cases.
\paragraph{Number of occurrences of a character (15 features)} We count the number of occurrences for characters: '\_', '-', '?', '!', ',', '.', '@', '\#', '\%', '\&', '=', '+', ';', ':', '/' yielding one feature for each character. For example, the occurrence counts indicate the presence of the fragment ('\#'), query ('?') or email in the URL ('@'), number of query ('\&', '=') and path ('/') parameters, amount of percent-encoded characters in the URL ('\%'), etc.
\paragraph{Number of non-Base64 characters (1 feature)} We count the number of characters in $s$ that are not part of the Base64 encoding. The feature can be used to easily recognize and separate URLs without Base64 encoding.
\paragraph{Non letter ratio (1 feature)} Given as 
\begin{equation}
r_{\text{non-}\alpha}(s) = 1-\frac{\text{number of letters in }s}{l(s)}.
\end{equation}
Obviously, this feature differentiates between alphabetic and non-alphabetic strings. 

\subsection{Trigram Based Features}
\label{sec:feat-3g}
We use probability distribution of trigrams not only in the context of DGA \cite{Antonakakis2012,Schiavoni2013}, but also to capture statistical properties of patterns/encodings in the path and query strings.

\paragraph{Single trigram based feature} Given a word dictionary, each word is decomposed to 3-grams (e.g.\ for \textit{abcd} we get two 3-grams: \textit{abc} and \textit{bcd}), and frequency of each unique 3-gram is computed as number of its occurrences divided by the number of all 3-grams in the input dictionary. 

\paragraph{Feature derived from multiple trigrams} The feature is given as the average: $n_Q(g_1) = (n(g_1) + n(g_2) + n(g_Q))/Q$, where $g_1, \ldots, g_Q$ are $Q$ adjacent 3-grams. Obviously, this is a smoothed version of the single trigram frequency feature suppressing the influence of peaks in the ranking of adjacent trigrams.

\paragraph{Feature vector construction} To get a feature vector of fixed length for a given string $s$ with variable length $l(s)$, $s$ is decomposed to 3-grams $g_1, \ldots, g_{l(s)-2}$, for each $g_i$ we compute $n(g_i)$ and construct a histogram with fixed number of bins given all the $n(g_i)$, $i = 1, \ldots, l(s)-2$. The final feature vector is given as concatenation of values in individual bins of the histogram. Feature vector for $n_Q(g)$ is constructed in the same way.

Intuitively, the resulting feature vector represents the distribution of the substrings (3-grams) in the input string indicating how common or rare they are. The notion of common and rare strongly depends on the choice of the word dictionary, which is task specific, see Section~\ref{sec:featsets}.

\subsection{Additional Features}
\label{sec:feat-flow}

Besides the previous features, we extract 12 additional proxy log based features. These are: flow duration, number of bytes transferred from client to server, client and server ports, number of bytes transferred from server to client, length of the user agent, length of MIME-type, HTTP status, and:

\paragraph{Repetitive changes of '$=$' and '$\&$' in URL} URL typically consist of repetitive changes of '$=$' and '$\&$' used to fill and separate individual fields of the corresponding form. This binary feature indicates, whether these two characters are repeated in the specified order, thus if the URL is in the ordinary form.
\paragraph{URL starts with specific string} We extract 3 binary features describing whether URL starts with "http://", "https://", "connect://", i.e. flow is ordinary HTTP request, encrypted HTTP request, specific HTTP tunnel to a remote end-point, respectively.

\section{Experimental Setup}

\subsection{Datasets}
During the system development and testing we used several datasets collected between November 2013 and April 2015 from enterprise networks as summarized in Table~\ref{tab:dataset}. The datasets were provided by the security company XYZ\footnote{The name of the company was removed to ensure anonymity of authors during the review process.}.
They contain diverse HTTP live traffic from proxy logs from 100+ companies from all over the world and represent, to the best of our knowledge, the largest datasets of this kind ever used in a published work. 

\begin{table}
\centering
\caption{Summary of used datasets.}
\begin{tabular}{|l|cccc|}
\hline
dataset ID & \#comp & \#users & \#flows & days \\ \hline
Nov2013 & 12 & 60k+ & 0.5B & 2 \\
Jan2014 & 70+ & 200k+ & 8.5B & 14 \\
Sep2014 & 50+ & 200k+ & 10.7B & 8 \\
Oct2014 & 60+ & 250k+ & 49B & 42 \\
Jan2015 & 70+ & 300k+ & 10.8B & 7 \\
Mar2015 & 100+ & 350k+ & 123B & 42
\\ \hline
summary & 100+ & 350k+ & 202.5B & 115
\\ \hline
\end{tabular}
\label{tab:dataset}
\end{table}

\subsection{Feature Sets}
\label{sec:featsets}
We use several feature sets:
%
\paragraph{Second Level Domain (SLD) set} We use histograms with 16 bins of both trigram-based features described in Section~\ref{sec:feat-3g} (for $n_Q(g)$ we use $Q=2$). Rather than using word dictionaries, which do not contain specific abbreviations and words that are commonly used in domain names, the probabilities for the 3-grams are computed from 1 million most popular sites obtained from Alexa\footnote{\url{http://www.alexa.com/}}. Also, the 26 features from Section~\ref{sec:feat-url} are extracted for SLD. Overall, we get $16 \times 2 + 26 = 58$ dimensional feature vector.
\paragraph{Path\&query set} Two histograms with 16 bins are now used for both path and query in the same way as for the SLD set yielding $4 \times 16 = 64$ features. Path and queries longer than characters are collected from the dataset Sep2014 and used to compute the probabilities for the 3-grams. Also, the 26 features from Section~\ref{sec:feat-url} are extracted for the path and query, i.e.\ $2 \times 26$ additional features. Overall, we get $64 + 2 \times 26 = 116$ dimensional feature vector.
\paragraph{Full set} Composed from the SLD and the path\&query set. Moreover, we extract the 26 features for the referrer, hostname, top-level domain, filename, i.e.\ $4 \times 26$ features for the four groups, and 10 additional features from the proxy log fields related to the uploaded/downloaded bytes, MIME type, HTTP status and the rest of features defined in Section~\ref{sec:feat-flow}. Overall, we get $58 + 116 + 4 \times 26 + 12 = 290$ dimensional feature vector. 
\paragraph{No-domain set} It is the full feature set without the features representing the domain, which we believe are not informative except for the malware with DGA. We extract only two domain based features: the number of occurrences of a dot (i.e.\ the number of sub-domains) and a dash, which is more common between malware sites. That is, the number of extracted features is $290 - 58 + 2 = 234$.
\paragraph{Pre-filtering set} This set is used by the zeroth pre-filtering level (see Fig.~\ref{fig:sys-full}), in order to decrease the computation complexity and the number of processed flows. It contains simple features related to the length of the referrer, and individual parts of the URL (see Fig.~\ref{fig:url-decomposition}) plus all features taken directly from the proxy log fields and the "repetitive changes of '$=$' and '$\&$'" feature (Section~\ref{sec:feat-flow}) plus 15 features related to the number of occurrences of special characters in the URL (Section~\ref{sec:feat-url}).

\subsection{Random Forest Hyperparameters}

Optimizing hyperparameters for each base classifier separately would be complicated, because we do not have a standard labeled training set. Labeled data are collected continuously also during the training process. Therefore, we performed 10 fold cross-validation in advance on Nov2013 from all the available labeled samples. Best values in terms of the precision and recall for the hyperparameters specified in Section~\ref{sec:rf} are: the number of trees $T=40$, the maximum depth of a tree $D_\text{max} = 19$, and the number of features $F$ selected randomly at a split node during the training process equals 1/3 of used features. These parameters were used for each base classifier.

\subsection{Evaluation}
\label{sec:eval}
We report the number of true positives (TPs) as the malware detected correctly and the number of false positives (FPs) indicating legitimate communication classified as malware.  
The FP-rate is defined as the number of FPs divided by the overall number of negatives. There are two major challenges when using the FP-rate to report IDS performance: a) accurately estimating the number of entities without any infection in a given dataset, b) deceivingly low value of the FP-rate in a large network with thousands of devices and hundreds of incorrect detections. The first issue is sometimes addressed by assuming that a network is well maintained, and therefore it has very few infections \cite{BotHunter,BotMiner,BotZilla,Perdisci2010}.

The TP-rate (accuracy or recall) is defined as the number of TPs divided by the total number of compromised devices. Therefore, the same challenges still hold. Ideally, both challenges are solved by using a fully labeled dataset which is not feasible in practical experiments. For these reasons, our primary indicator of the IDS performance is the \textit{precision} defined as the number of TPs divided by the total number of all detections, i.e.\ TPs+FPs.

To further illustrate the drawbacks of the FP-rate in our setting, let us assume that all users in Mar2015 dataset without detected incidents are clean. For example, in the case of DGAs, we have 134 FP and 1234 TP users, see Table~\ref{tab:res-prod}. The FP-rate would be computed as $\#FPs/\#clean\_users = 134/(350 \times 10^3-1234) = {3.84 \times 10^{-4}}$. Now, if we assume that 50\% of the devices are infected, the FP-rate is $134/(175 \times 10^3-1234) = 7.71 \times 10^{-4}$. Clearly, the FP rates are very low (and in the same order of magnitude) in both cases despite the dramatic difference in the number of infections. Therefore, we report the results in terms of precision, much more informative measure in our setting.

When deploying a detection system, the most relevant information is the precision on the user level, i.e.\ the percentage of truly compromised users out of all the users detected as compromised. Our system identifies the compromised users and malicious domains as soon as one malicious flow is detected. Clearly, there can be many correctly detected malicious flows but all having the same domain name. Similarly, one malicious domain can affect multiple users while several malicious domains only a single user (as is the case for DGA, see Section~\ref{sec:dga-deploy}).
Therefore, we will report the results in terms of the detected proxy logs, unique URLs, domains, and users. 
The proxy logs can be grouped according to URLs, domains, or users. For example, to compute the precision on the user level, all the proxy logs of a user over the whole evaluation period are collected, and the user classification is reported as a TP if at least one of the proxy logs was detected as malicious. In addition, if the user's proxy logs contain logs which were classified as malicious, but are legitimate, the classification of the same user is also reported as an FP. The precision for URLs and domains can be obtained similarly. However, each user, domain, or URL is reported as a TP/FP at most once during the whole evaluation period.

\textit{The labeling was done on the domain level}, i.e.\ if a domain was found to be malicious, all the flows to this domain were also assumed to be malicious. To verify the maliciousness we used conventional tools such as VirusTotal\footnote{http://www.virustotal.com}, page rank, and additional internal malware feeds. Since the investigation of maliciousness is hard for some traffic, we usually report it as an FP in cases when there is no clear evidence that the communication is malicious. Therefore, the reported results can be seen as a lower bound on the precision.

We also report the Receiver Operating Characteristics (ROC) along with other evaluation statistics, all summarized in Fig.~\ref{fig:roc}. The analysis and labeling of the detection output for the whole range of the detection threshold $0 \leq \tau < 1$ would be very expensive, because the number of detections for $\tau = 0$ is in order of magnitude higher than for $\tau = 0.5$. Therefore, we report only operating characteristics for $\tau > 0.5$.

In order to analyze the impact of our system, we compared the results to the industry-leading signature and rule-based IDS (SRB-IDS), already deployed and operating on the same data. \lmanew{They are based on majority of available feeds, URL regexp patterns, domain/IP blacklist, and global and local reputation systems.}
We report the amount of blocked proxy logs, URLs, hostnames, and users relative to all the detections of the multi-level system.

\section{Malware Classification}
\label{sec:mal-class}
The idea is to decompose the behavior to the characteristic patterns represented by the base classifiers, and focus on malware specific manifestations at the last classification level. We do not aim to detect all the flows from a particular malware communication, we aim only for a subset of flows when the malware exposes itself in terms of distinguishing from the legitimate communication.

Next sections contain detailed description of the building process of individual detectors, the collection of training data, and the design of each (base) classifier. At the end of each section, we report evaluation results after the detectors were deployed. In Section~\ref{sec:ENC} we will describe how to build a multi-level system for the detection of the malicious traffic using encrypted strings. In the last part of this section, we will also demonstrate how to use the architecture for training a robust detector of malware using a DGA.

\begin{figure}
\centering
\includegraphics[width=0.4\linewidth]{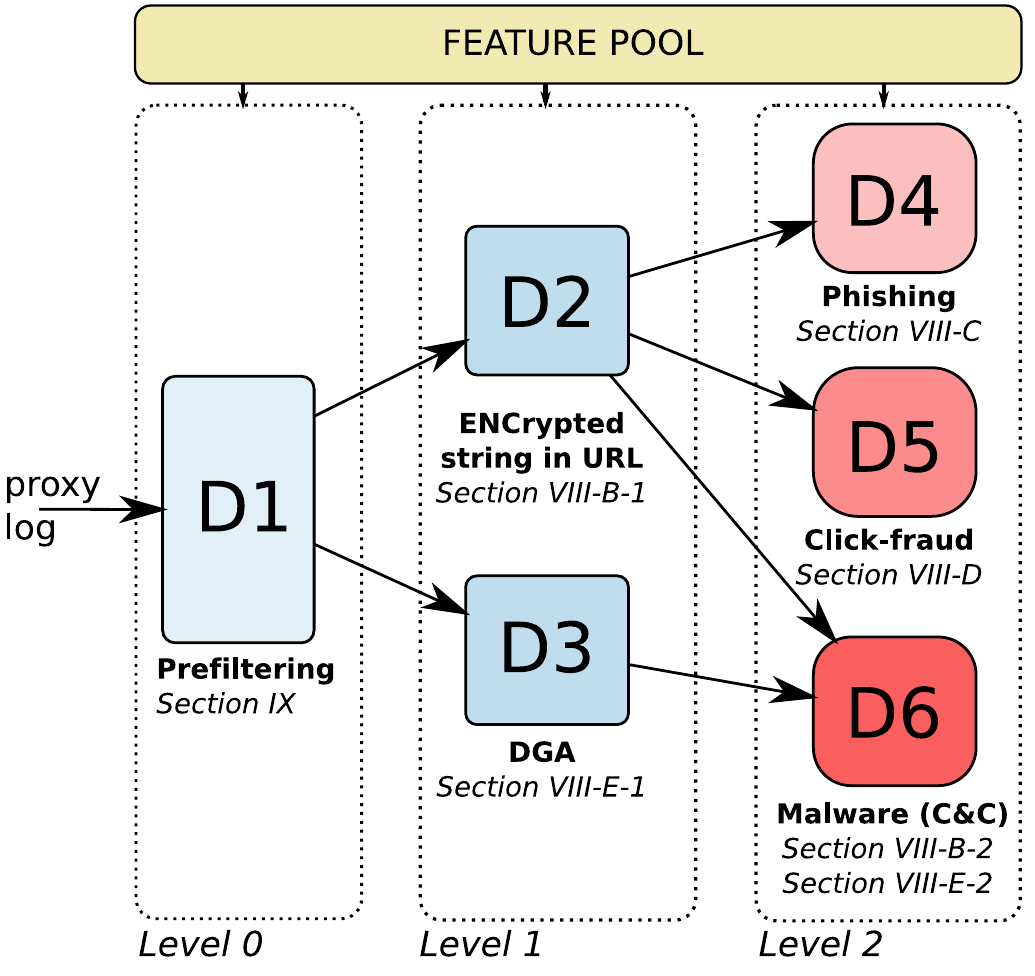}
\caption{Overview of the complete detection system with two detection levels and a zeroth pre-filtering layer used to reduce the amount of data processed by subsequent levels.}
\label{fig:sys-full}
\end{figure}

\subsection{Detection Workflow}
\label{sec:workflow}
The overview of the entire multi-level system, whose components will be described in the following sections, is depicted in Fig.~\ref{fig:sys-full}. A particular detection path represents one multi-level classifier discussed in Section~\ref{sec:arch}. The input of the system is a single flow (proxy log). Each level consists of different detectors, provided by base-classifiers. At each level, features requested by a given detector are extracted from the proxy log fields yielding a feature vector, which is processed by the base classifier. First level contains two detectors identifying (not necessarily malicious) generated domains and encrypted strings in the URL. For each detector a decision score $s$ is computed and compared with the decision threshold $\tau$. If $s > \tau$ for at least one of the detectors, the proxy log is passed to the next level according to paths specified in Fig.~\ref{fig:sys-full}. For example, in the case of the encrypted strings, we are able to distinguish phishing, click-fraud, and general C\&C communication patterns. An incident is created only if one of the level-2 detectors triggers an alert, i.e.\ the detection score of base classifiers at the second level is higher than $\tau$. Detections can be prioritized according to the type of the infection, e.g.\ malware (C\&C) has highest priority while phishing lowest.

The operating point of the trained detectors for which the results will be reported was set by using a threshold $\tau = 0.5$. Therefore, more than 50\% of the trees in the random forest have to vote for the sample to be detected by a particular base classifier. Detailed evaluation statistics for the full range $0.5 < \tau < 1$ can be found in Fig.~\ref{fig:roc}.

To reduce the load of the feature extraction (see Section~\ref{sec:costs}), we use the zeroth level, block D1 in Fig.~\ref{fig:sys-full}, with the simple pre-filtering feature set defined in Section~\ref{sec:featsets}. The zeroth level is trained after all the other levels were trained and it is retrained when a new detector is added. The operating point is set to 100\% recall given all the evaluation samples. In this way, we are able to filter out approximately 60\% of incoming flows at level-0 with regular URL patterns, and in this way decrease the computation costs.

\subsection{Encrypted Strings in URL (C\&C)}
\label{sec:ENC}
Our detection system is based on a two-level classifier build in a bottom-up fashion.
We are interested in communication with long URLs with most of the path and query encrypted, therefore 
URLs with only a few encryptions, mostly of some query keys, are regarded as false positives. Examples are given in Table~\ref{tab:tunn-ex}.
Summary of training parameters can be found in Table~\ref{tab:DGA-trsets}.

\begin{table} 
\centering
\caption{Summary of samples in positive and negative training sets, type of the feature set and hyperparameters setup for individual levels and after retraining. Specification of hyperparameters is in the form ($T$, $D_\text{max}$, $F$), see Section~\ref{sec:rf}. Note that $F$ equals approximately 1/3 of the size of the respective feature set.}
\begin{tabular}{|l|cccc|}
\hline
& \#pos & \#neg & feat-set & rf-params \\ \hline
DGA-L1 & 1M & 51k & SLD & (40, 19, 20) \\
DGA-L2 & 39k & 28k & full & (40, 19, 100) \\
DGA-L2-RE & 40k & 35k & full & (40, 19, 100)
\\ \hline
ENC-L1 & 30k & 99k & path\&query & (40, 19, 40) \\
ENC-L2 & 31k & 31k & no-dom & (40, 19, 80) \\
ENC-L2-RE & 52k & 52k & no-dom & (40, 19, 80)
\\ \hline
PHISH & 20k & 104k & no-dom & (40, 19, 80)
\\ \hline
CLICK-FRAUD-RE & 35k & 130k & no-dom & (40, 19, 80)
\\ \hline
\end{tabular}
\label{tab:DGA-trsets}
\end{table}

\subsubsection{Training ENC-L1} 
This detector represents the block D2 in Fig.~\ref{fig:sys-full}. First level uses only features extracted from path and query strings of the URL. Note that we utilize also the trigram based features typically used for detection of DGAs. To get the negative sample set, we made use of Alexa and the first 500k most popular domains. These were searched for in Jan2014. Since a large amount of features is extracted from the URL, only one proxy log was taken per each unique URL in order to avoid overfitting. It was chosen randomly from all available proxy logs with the same URL. For each found domain we kept the first proxy logs (on a unique URL basis). These were then processed by simple hand crafted rules to discover encryptions: e.g.\ the URL is first split to strings according to common symbols (e.g. \&, /), and the rules are based on the length of the parts after split, ratio of upper and lower case letters or ratio of numbers is computed. Results were then manually inspected and minor corrections were made. Proxy logs with long URLs having most of the path/query encrypted were discarded from the training since we would like to classify these at this level as encrypted URLs, they also serve as the negative sample set for the training of ENC-L2. We ended up with a negative training set comprising 29k of unique domains and 99k of respective proxy logs. 

In order to form the positive set we took all malware samples from all the available sources: blacklists, sandboxing reports and DGA samples caught by our DGA detector discussed in Section~\ref{sec:dga-tr}. Proxy logs with these domains were searched for in Jan2014. Only unique URLs were kept and manual inspection along with rules described above were used to discard those without encryption. We collected proxy logs with encrypted URLs from 225 different malicious domains yielding 31k of proxy logs.

Both sets were used to train the ENC-L1 detection level.

\subsubsection{Training ENC-L2}
This detector is part of the block D6 in Fig.~\ref{fig:sys-full}. The malicious encrypted flows from ENC-L1 were used as the positive training set. To get the negative set, we run ENC-L1 on Jan2014. The flows that passed the detection and the domain of the flow appeared in the first 500k of domains in Alexa, were used as the negative set (5477 domains with 31k of proxy logs). The ENC-L2 classifier was trained utilizing the full feature set, i.e. full information content, see Table~\ref{tab:DGA-trsets}.

\begin{table}
\centering
\caption{Performance of the 2-level encryption detector on Jan2015 after retraining.}
\begin{tabular}{|l|cccc|}
\hline
Jan2015 &  logs &  unq-URLs  &  domains  &  users 
\\ \hline
TPs [count]  & 25,234 & 22,405 & 398 & 1131
\\
FPs [count]  & 2917 & 2761 & 69 & 165
\\ \hline
prec [\%]  &  89.64 &  89.02  &  85.22  &  87.27 
\\ \hline
\end{tabular}
\label{tab:ENC-L2-jan}
\end{table}

\paragraph{Retraining ENC-L2}
Given the two level detector, we performed several rounds of retraining on Jan2015 dataset. Since the first level is responsible for the detection of encryptions, just the second level responsible for detection of C\&C traffic was retrained in each round. In each iteration, output of the detector was searched for TP domains, flows to these domains were added to the positive set (on a unique URL basis), and the rest of flows enlarged the negative set. The system was retrained and the performance was evaluated on Jan2015. The process was stopped when the precision on the user level reached 85\%, which happened already after one iteration. We ended up with training sets containing 52k negative and 52k positive samples (proxy logs). Summary of the training parameters can be found in Table~\ref{tab:DGA-trsets}. Results after the retraining are reported in Table~\ref{tab:ENC-L2-jan}.

\subsubsection{System Deployment}
\label{sec:dga-deploy}
\begin{figure}[t!]
\centering
	\begin{subfigure}[b]{0.49\textwidth}
	  \includegraphics[width=\textwidth]{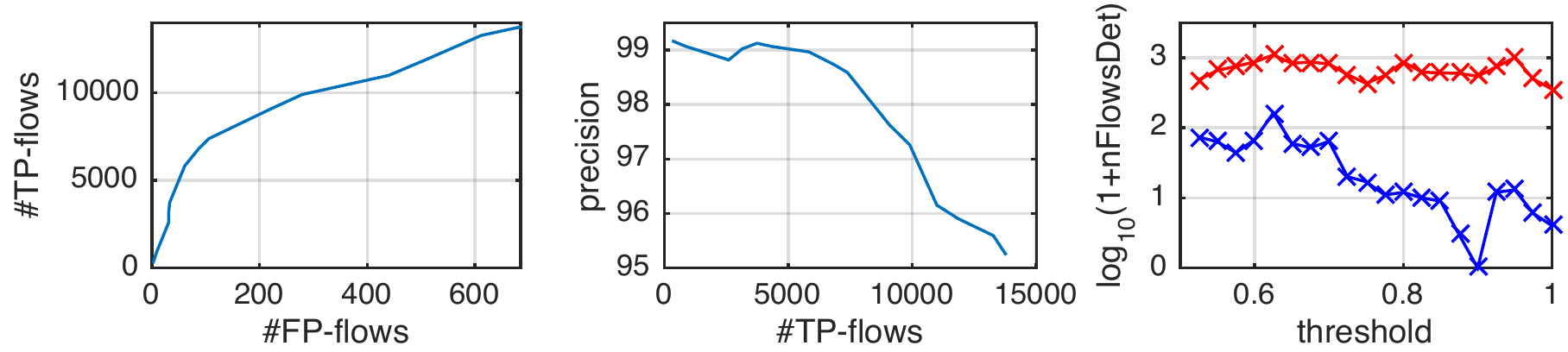}
	  \caption{ENCryption detector evaluated on Mar2015.}
 	\end{subfigure}
	\begin{subfigure}[b]{0.49\textwidth}
	  \includegraphics[width=\textwidth]{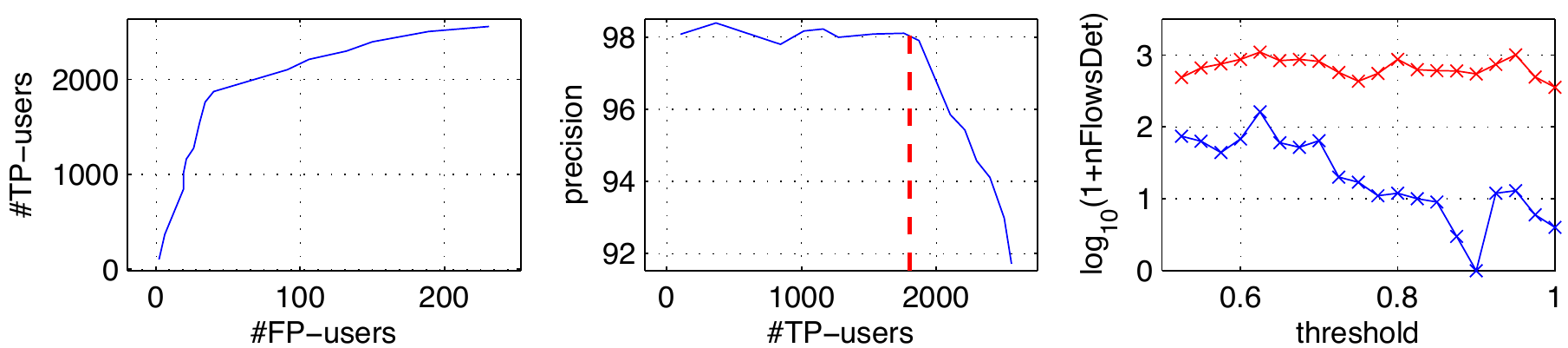}
	  \caption{ENCryption detector evaluated on Mar2015.}
 	\end{subfigure} 	
 	\\
	\begin{subfigure}[b]{0.49\textwidth}
	  \includegraphics[width=\textwidth]{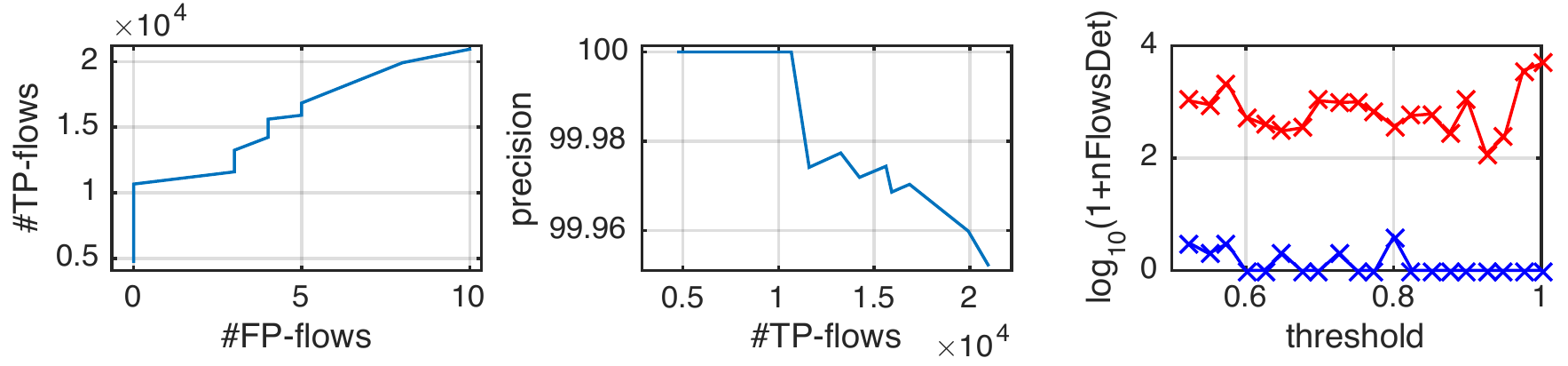}
	  \caption{Click-fraud detector evaluated on Mar2015.}
 	\end{subfigure}
	\begin{subfigure}[b]{0.49\textwidth}
	  \includegraphics[width=\textwidth]{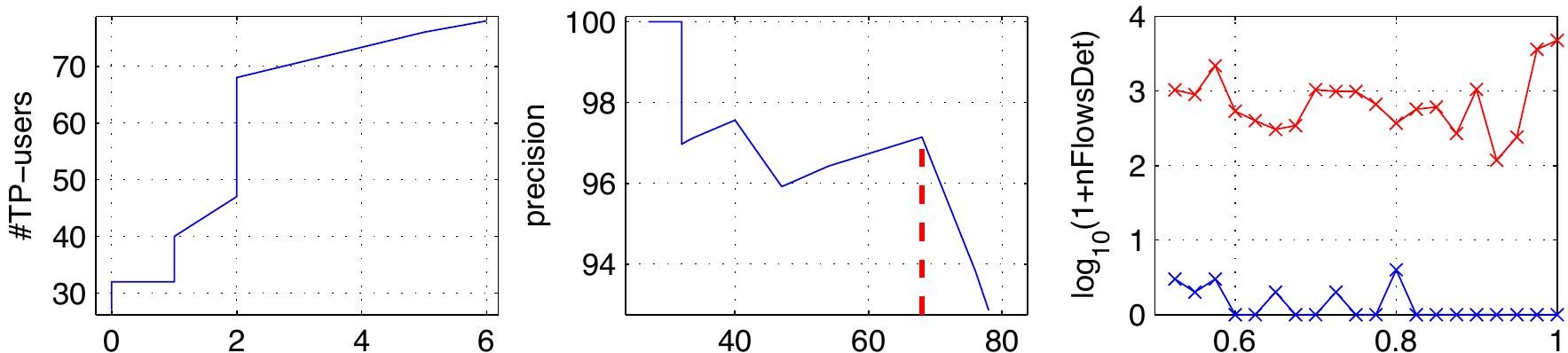}
	  \caption{Click-fraud detector evaluated on Mar2015.}
 	\end{subfigure} 	
 	\\
	\begin{subfigure}[b]{0.49\textwidth}
	  \includegraphics[width=\textwidth]{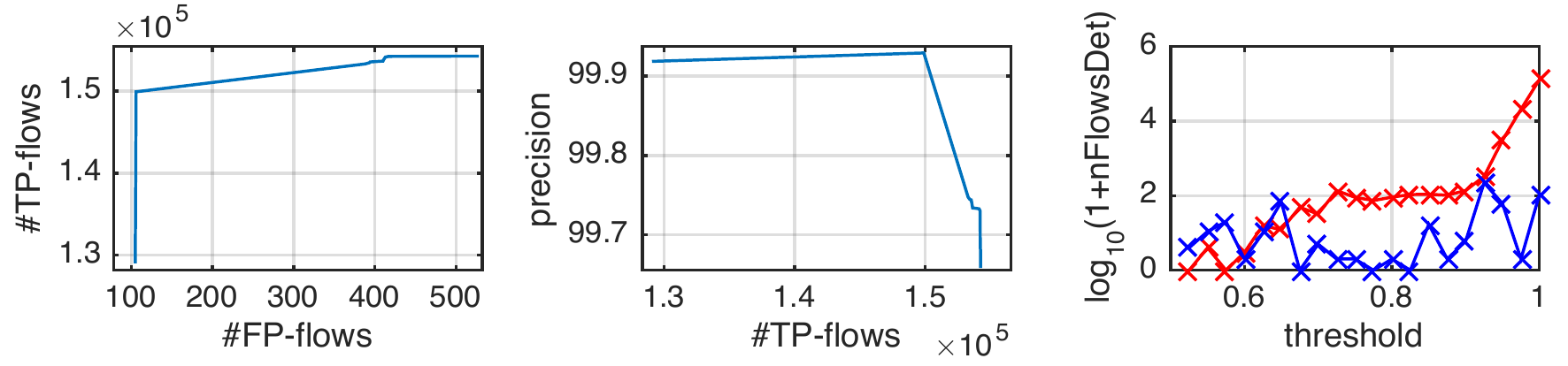}
	  \caption{DGA detector evaluated on Mar2015.}
 	\end{subfigure}	
	\begin{subfigure}[b]{0.49\textwidth}
	  \includegraphics[width=\textwidth]{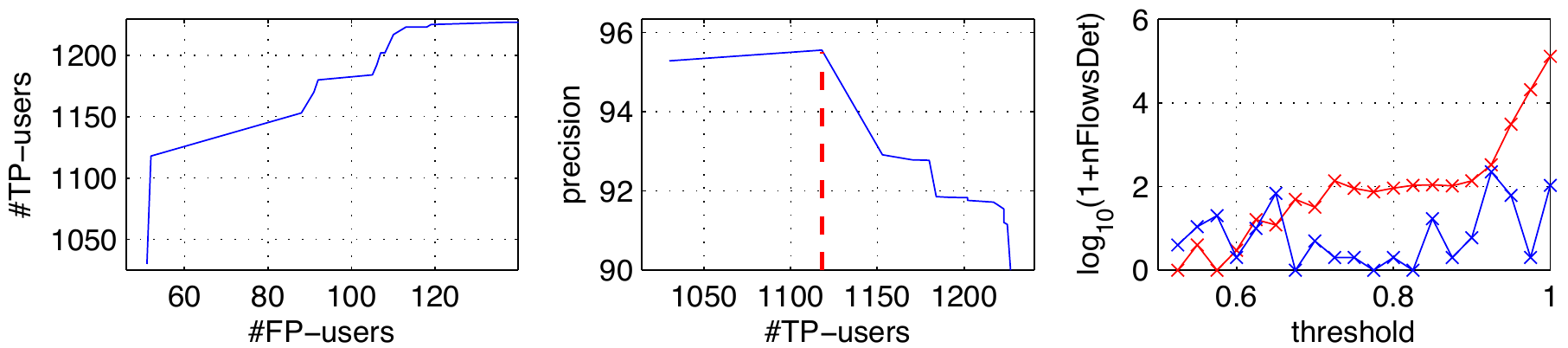}
	  \caption{DGA detector evaluated on Mar2015.}
 	\end{subfigure} 	 
 	\caption{Results given separately for the flows (sub-figures (a), (c), (e)) and users (sub-figures (b), (d), (f)). For each of the multi-level classifier (rows), we varied the detection threshold to obtain ROC curves (1st column), precision in dependence on the number of TP flows/users (2nd column), and the number of detected flows/users (3rd column; red = malicious, blue = legitimate). Changing the threshold between $0.5$ and $1$ moves the operating point between the right-most and the left-most point on the curves, respectively. Phishing was left out of the plots since the precision is already at 100\% for $\tau = 0.5$.
 	The precision can be obviously further improved by increasing the threshold beyond the working point of $\tau = 0.5$. 
 	For DGA and ENC classifier, we cannot reach precision 100\%, because the score for the legitimate flows/users is distributed along the full threshold range, see 3rd column of graphs.
	Note that the precision on the user level can be further improved by increasing the threshold beyond the working point of $\tau = 0.5$. This is depicted by the dashed red line in the second column. For the DGA classifier we can arrive at a precision above 95\% while keeping recall above $1120/1234 \times 100 \approx 90\%$ (red dashed line). In the case of the encryption detector ENC, increasing the detection threshold value will increase the precision up to 98\% with recall still above $1870/2550 \times 100 \approx 73\%$.}
 	\label{fig:roc}
\end{figure}

We analyzed the output of the system running in production environment on a weekly basis over 6 weeks without any additional retraining. Results on Mar2015 are given in Table~\ref{tab:res-prod}, Fig.~\ref{fig:roc}.
The performance of the system exceeds 95\% on the flow level and 90\% on the user level, out of which only 9.69\% of infected users were detected also by SRB-IDS, i.e. we deliver a substantial improvement in detection of a significant malware pattern based on encrypted strings in the URL. On this dataset, ENC (C\&C) reports 2550 TPs + 256 FPs = 2806 detections in total. This yields $2806 / 100 companies \approx 28$ detections per company per month, out of which at most 3 users are false positives. This is an acceptable amount for a deployed advanced threat detection system. Based on Fig.~\ref{fig:roc}, we can further increase the user precision up to 98\% however loosing 27\% of TP users.

We also analyzed the false positives. They are related mainly to non-standard encodings of strings in the URL's path and query, used for example to transfer strings with unsupported characters. Another class of FPs is related to unconventional URL shortening services.

\begin{table}
\centering
\caption{Performance of the deployed system for individual detectors on the Mar2015 dataset at working point $\tau = 0.5$. We report also the amount of TPs detected by our system as well as by the signature and rule-based IDS (SRB-IDS), already deployed and operating on the same data. Note that the precision is above 90\% in all the cases, i.e.\ one of ten detected users is a false positive. The precision can be further boosted tightening the decision threshold, see Fig.~\ref{fig:roc}.}
\begin{tabular}{|l|cccc|}
\hline
 &  proxy logs  &  unq-URLs  &  domains  &  users 
\\ \hline \hline
\textbf{ENC (C\&C) Mar2015}  &  \multicolumn{4}{|l|}{\textbf{\textit{Signature and rule-based technologies}}} \\ \hline
TPs [count]  & 1044 & 633 & 154 & 247
\\
FPs [count]  & 132 & 80 & 17 & 20
\\ \hline
precision [\%]  &  88.78  &  88.78  &  90.01  &  92.51 
\\ \hline
 &  \multicolumn{4}{|l|}{\textbf{\textit{Multi-level ENC (C\&C) detector}}} \\ \hline
TPs [count]  & 13,761 & 10,810 & 630 & 2550
\\
FPs [count]  & 685 & 503 & 119 & 256
\\ \hline
precision [\%]  &  \textbf{95.26}  &  95.55  &  84.11  &  90.88
\\
detected by SRB-IDS [\%]  &  \textit{\textbf{7.59}} &  5.86  &  24.44  &  9.69
\\ \hline \hline
\textbf{PHISH Mar2015}  &  \multicolumn{4}{|l|}{\textbf{\textit{Signature and rule-based technologies}}} \\ \hline
TPs [count]  & 8842 & 7885 & 42 & 617
\\
FPs [count]  & 0 & 0 & 0 & 0
\\ \hline
precision [\%]  &  100.0 &  100.0  &  100.0  &  100.0 
\\ \hline
 &  \multicolumn{4}{|l|}{\textbf{\textit{Multi-level phishing detector}}} \\ \hline
TPs [count]  & 19,666 & 17,561 & 58 & 1102
\\
FPs [count]  & 0 & 0 & 0 & 0
\\ \hline
precision [\%]  &  \textbf{100.0}  &  100.0  &  100.0  &  100.0
\\
detected by SBT [\%] & \textit{\textbf{44.96}} &  44.90  &  72.41  &  55.99
\\ \hline \hline
\textbf{CLICK-FR Mar2015}  &  \multicolumn{4}{|l|}{\textbf{\textit{Signature and rule-based technologies}}} \\ \hline
TPs [count]  & 5,304 & 5,304 & 8 & 2
\\
FPs [count]  & 0 & 0 & 0 & 0
\\ \hline
prec [\%]  &  100.0 &  100.0  &  100.0  &  100.0 
\\ \hline
 &  \multicolumn{4}{|l|}{\textbf{\textit{Multi-level click-fraud detector}}} \\ \hline
TPs [count]  & 20,949 & 22,069 & 36 & 78
\\
FPs [count]  & 10 & 8 & 4 & 6
\\ \hline
precision [\%]  &  \textbf{99.95}  &  99.96  &  90.00  &  92.86
\\
detected by SRB-IDS [\%]  &  \textit{\textbf{24.01}} &  24.03  &  22.22  & 2.56
\\ \hline \hline
\textbf{DGA Mar2015}  &  \multicolumn{4}{|l|}{\textbf{\textit{Signature and rule-based technologies}}} \\ \hline
TPs [count]  & 143,498 & 12,649 & 3718 & 284
\\
FPs [count]  & 23 & 19 & 15 & 17
\\ \hline
precision [\%]  &  99.98  &  99.85  &  99.60  &  94.35 
\\
\hline
 &  \multicolumn{4}{|l|}{\textbf{\textit{Multi-level DGA detector}}} \\ \hline
TPs [count]  & 154,249 & 14,656 & 4705 & 1234
\\
FPs [count]  & 526 & 322 & 64 & 134
\\ \hline
precision [\%]  &  \textbf{99.66}  &  97.85  &  98.66  &  90.20
\\
detected by SRB-IDS [\%]  &  \textit{\textbf{92.99}} &  86.31  &  79.02  &  23.01 
\\ \hline 
\end{tabular}
\label{tab:res-prod}
\end{table}

\subsection{Phishing}
\label{sec:phish-exp}
The analysis of the output of the ENC-L1 encryption detector revealed 10 domains that were later linked to phishing (see example URLs in Table~\ref{tab:phish-ex}). We collected 20k flows (out of which 80\% was blocked by SRB-IDS) for these 10 domains from the Sep2014 dataset. These form the positive training set. Since the communication pattern is based on the encrypted strings, the first level is shared with the ENC classifier. The phishing classifier represents the block D4 in Fig.~\ref{fig:sys-full}. The training setup of the second level phishing classifier was the same as for ENC-L2, see Table~\ref{tab:DGA-trsets}. The positive and negative samples from the training set of ENC-L2 were used for as the negative training set for this level. After the classifier was trained it was evaluated on Mar2015. The results are reported in Table~\ref{tab:res-prod}. 

\begin{table}
\centering
\caption{Example of the phishing-related communication.}
\begin{tabular}{l}
\hline
\textbf{malicious} \\
\hline
\textit{hxxp:}//cohostro[.]com/keyJjIjogMjM4MiwgImYiOiAwLCAibSI6IDI2M..
\\ 
\textit{hxxp:}//coclouderx[.]com/peyJjIjogMjUyOCwgImYiOiAwLCAibSI6IDI3..
\\
\textit{hxxp:}//adminxhost.com/8a766814986811e486083c4a92db07ce
\\ \hline
\end{tabular}
\label{tab:phish-ex}
\end{table}

The detector could learn the pattern and separate it perfectly from the rest of the traffic since the precision is at 100\%. However, 60\% of the domains were already seen during the training on the Sep2014 dataset, but the flows/URLs were different. In summary, we were able to build a new highly precise detector for a specific type of behavior, reusing already trained components without any additional effort. It is remarkable, that the statistical features employed during learning reliably represent the encrypted strings which allows the two behaviors to be separated even without the knowledge of the encryption algorithm. Note that 72\% of detected domains were detected also by SRB-IDS, but on the flow level only 45\% of flows was detected covering 56\% of users.

\subsection{Click Fraud}
\label{sec:clickfraud}

\begin{table}
\centering
\caption{Example of the click-fraud-related communication.}
\begin{tabular}{l}
\hline
\textbf{malicious} \\
\hline
\textit{hxxp:}//findreek[.]com/cen?ag=de6b7ffbf8a767e8bdecbb805143bca6-2..
\\
\textit{hxxp:}//199.182.165[.]105/c.php?i=DFuwjKoDUiUNzF8Qnn\%2F\%2FQw..
\\
\textit{hxxp:}//clickered[.]com/cex?si=94b2ba8b0b59787b4609c64514baa26c-81-0
\\
\textit{hxxp:}//180.149.131[.]33/v.php?q=252043836\&callback=jQuery11020...
\\
\textit{hxxp:}//lookfunnel[.]com/lr?si=362edf9becb1bf79713d1cf131936afb-18-0
\\ \hline
\end{tabular}
\label{tab:clck-ex}
\end{table}

The analysis of the ENC-L1 detector also revealed a pattern related to click fraud (see examples in Table~\ref{tab:clck-ex}). We collected 45 domains with 35k of proxy logs in the Sep2014 dataset that were used as the positive training set of the new base classifier. Since the communication pattern is based on the encrypted strings, the first level is shared with the ENC classifier, see Fig.~\ref{fig:sys-full} -- block D5. The training setup of the second level click-fraud classifier was the same as for ENC-L2, see Table~\ref{tab:DGA-trsets}. All samples from training of the previous classifiers were used as the negative set, i.e.\ ENC-L2 positive and negative training samples and phishing positive training samples. The classifier was evaluated on Oct2014 and the identified FPs were used to retrain the classifier. After the retraining, it was evaluated on the unseen Mar2015 dataset. 

Results are reported in Table~\ref{tab:res-prod} and Fig.~\ref{fig:roc}. Out of 36 detected domains, only 5 domains were seen during the training (although with different URLs/flows). Similarly to phising, most of the proxy logs have unique URL as seen from Table~\ref{tab:res-prod}. Most likely, each URL carries specific user and click-related malware instructions. The disproportion between number of detected FP and TP flows is huge: 10 to 20,949. However, almost each FP flow is linked with a different user. The few FP flows were identified as being part of a marketing and tracking services using similar URL patterns, however we were not able to find any obvious association with malware.

\subsection{Malware using DGA}
\label{sec:dga-tr}
Here we show how the proposed architecture can be used to build a DGA detector based only on the data obtained from a proxy log. We focus on DGAs producing out-of-vocabulary words, the features are not designed for detecting word-based DGAs.
The detection of DGAs is analogical to the detection of encryptions since the dissimilarity between malicious and legitimate communication hinges on the distribution of characters in both cases.
We use the same architecture as for the detection of encryptions in the URL built in the bottom-up fashion. 
Summary of the training parameters can be found in Table~\ref{tab:DGA-trsets}.

\subsubsection{Training DGA-L1} 
This detector represents the block D3 in Fig.~\ref{fig:sys-full}. The first level uses features extracted from the second level domain only to filter out those domains that are clearly not generated by a DGA. 
The operating point of the first stage is therefore set for high recall (and lower precision). In order to train the first level, the system uses one million most trusted domains from Alexa to form the set of negative/legitimate samples. 

The set of positive/malicious samples is taken from various sources of DGA domains, such as blacklists, sandboxing reports, domain lists produced by new types of DGAs, and lists gathered by DNS anomaly detectors designed to analyze unusual activity in DNS requests \cite{AnomSurve2009,RehakPGB08}. Note that the DNS anomaly detectors are used only in the training process to collect training data. They  are tuned to operate with high precision at the cost of lower recall. This is to ensure that the domains are indeed reliable positive samples generated by a DGA. The various data intelligence sources are sometimes complementary with the goal to cover all possible variants of DGAs. Overall, 51k of malicious DGA domains were collected to form the positive training set. To get an estimate of the performance of DGA-L1, we run the classifier on Nov2013 (with detection threshold $\tau=0.5$) and obtained 594 domains. We could confirm maliciousness for 87 of them yielding lower bound for precision of DGA-L1 on the domain level equal to $87/594 \times 100 = 14.65\%$. Typical FPs for level-1 were described in Section~\ref{sec:dga-ex} and are given in Table~\ref{tab:dga-ex}. We run also a query on the Nov2013 dataset given the collected malware domain set, but we found only flows to 55 domains that were a subset of the detected 87 domains. Obviously, the DGA domains rotate quickly.

\subsubsection{Training DGA-L2}
Detector DGA-L2 is part of the block D6 in Fig.~\ref{fig:sys-full}. The second level uses the full feature set extracted from proxy logs to model the full malware behavior and to filter out false positives from the first stage. It is trained only from data that passed the detection on the first level. Only one proxy log, chosen randomly from all available proxy logs with the same URL, was taken per each unique URL. Otherwise, domains with majority of flows would be favored and the classification function would model primarily communication to these domains. Interestingly, number of proxy logs with unique URL to the malicious DGA domains was 9k, while number of proxy logs for remaining $594-87=507$ domains was 1k. 
Since we regarded these numbers still too low to train a robust classifier, we performed the following steps in order to extend the training set. 
\begin{enumerate}
\item We enlarged the DGA-L2 malicious training set by mixing-in the domain names from the first stage and augmenting them with the other proxy log fields in the training samples. The mixing is done by iteratively replacing the second level domains from the proxy logs by the new second level domain samples. Although this produces URLs to non-existing locations, the statistical properties of the samples are preserved. This step increases the variability of the proxy log samples, enlarges the training set, and makes the resulting classifier more robust.
Since large number of features is extracted from URL's path and query strings, reproducing the training sets altering the domain names only yields multiple occurrences of every path and query and thereby promote features related to path and query. Therefore, in the first mixing-in we exchanged only the domain names and kept all the other fields untouched, while in the second mixing-in we removed path and query from the URL. In such a way we tripled the number of malicious proxy logs mixing in additional 18k of randomly taken domains from the malicious DGA-L1 domain set.
\item We temporarily lowered the detection threshold for DGA-L1 to $\tau=0$, i.e. if one of the trees from DGA-L1 classified the sample as having generated domain, the sample was put to the DGA-L2 training set. The number of domains incresed from 507 to 12k with 39k of proxy logs with unique URL. They were all used as legitimate/negative sample set for the DGA-L2 without further analysis. This measure was taken in order to increase the precision of the second level at the cost of recall.
\end{enumerate}
Overall, we have got 39k of negative and 28k of positive proxy logs used as negative and positive training sets, respectively. Summary of the training parameters can be found in Table~\ref{tab:DGA-trsets}.

\paragraph{Retraining DGA-L2}
Given the two-level detector, we performed several rounds of retraining. In each round, only the second level was retrained. The first level is based only on the features extracted from the domain, therefore it cannot achieve high precision. At the same time, we wish to keep the recall at this level high. The retraining was done on the Oct2014 dataset. In each iteration, the output of the detector was searched for TP domains, the flows to these domains were added to the positive set, the rest of the flows to the negative set, and the system was retrained. The process stopped when the precision on the user level reached 85\% (2 iterations). The results are given in Table~\ref{tab:DGA-L2-retr}.

\begin{table}
\centering
\caption{Performance for the retrained DGA detector on Oct2014.}
\begin{tabular}{|l|cccc|}
\hline
\textbf{Oct2014}  &  \#logs  &  \#unq-URLs  &  \#domains  &  \#users 
\\ \hline
TPs  & 127,932 & 9884 & 1056 & 797
\\
FPs  & 695 & 112 & 45 & 114
\\ \hline
prec [\%]  &  99.46 &  98.88  &  95.91  &  87.49 
\\ \hline 
\end{tabular}
\label{tab:DGA-L2-retr}
\end{table}

\subsubsection{System Deployment}
Results of the system running over 6 weeks in the production environment without any additional retraining are given in Table~\ref{tab:res-prod}, Fig.~\ref{fig:roc}. Note that a malicious domain, but not generated, is from the DGA detection perspective a false positive.

According to Table~\ref{tab:res-prod}, the detector's precision is above 99\% on the proxy log level and 90\% on the user level with only 23\% of users detected also by SRB-IDS. This result shows that our system complements the signature and rule-based solutions by capturing additional threats. 
Moreover, according to Fig.~\ref{fig:roc}, the user precision can be boosted up to 95\% loosing approximately 10\% of TP users. 

Most of the FPs were related to flows to non-English domains. This is caused by the fact that in the training we used mainly the most visited domains from Alexa to form the negative set, which are formed mainly from common English words. Still, the number of detected FP flows was very low (526), but each flow was linked with a different user reducing the precision from 99\% on the flow level to 90\% on the user level. This could be solved for example by introducing an additional level between level-1 and level-2 (or replacing level-1), which would contain detectors for individual languages.

To get the idea about the drop in the number of detections, we can compare the results with Table~\ref{tab:DGA-L2-retr}, where more than 1000 malicious domains were detected, with precision on the user level at 87\%. Clearly, the number of detections on the larger dataset, Mar2015, is high even 5 months after the deployment without any retraining of the classifier. This can be attributed to the generalization performance of the detection system.

Note that in Table~\ref{tab:res-prod}, the number of flows is ten times higher than the number of unique URLs. Although the system is designed to detect malicious flows, the deployment is possible only when achieving high precision also at the user level. This is important since there may be many different DGA domains in a communication of several users. Handful of false-positive DGA detection spread across many users might cause lower precision at the user level while maintaining high precision at the proxy log level.

\section{Performance Analysis}
\label{sec:costs}
Using random forests described in Section~\ref{sec:rf} with $T$ trees, each tree of maximal depth $D_\text{max}$ and $L$ levels of hierarchy in the multi-level classifier, the upper bound on the number of operations (i.e.\ binary tests at individual nodes of the decision tree) to get the detection result is $N = T \times D_\text{max} \times L$. That is, the detection complexity is $O(N)$, i.e.\ linear in the number of tests performed using features for each proxy log. In practice, the number of comparisons is much lower, since many samples are filtered out at the lower levels or the traversed path in a tree has much lower depth than $D_\text{max}$.

To approximate the number of operations in feature extraction, note that 25 of the string features from Section~\ref{sec:feat-url} plus one feature for repetitive changes defined in Section~\ref{sec:feat-flow} can be extracted in a single pass through the characters in the input string. That is, the number of operations is lower bounded by the length of the input string. Extracting these features both for the URL and referrer of lengths $S^\text{url}$ and $S^\text{ref}$, respectively, yields $(26 \times S^\text{url} + 26 \times S^\text{ref})$ operations. 
Extracting the histograms of trigram frequencies (defined in Section~\ref{sec:feat-3g}) from a URL string can be done by one hash map lookup for a specific trigram in the string and another hash map lookup to update a specific bin in the histogram of frequencies. Together, given a URL of length $S^\text{url}$, the lower bound on the number of operations is $2 \times (S^\text{url}-2)$, where $(S^\text{url}-2)$ is the number of trigrams.
We do not count any operation for extraction of features that are taken directly from the proxy log fields (e.g.\ http-status, transfered bytes, port number; see Section~\ref{sec:feat-flow}), we count one operation for features related to the string lengths (i.e.\ 6 operations in total), and at most 25 operations (one operation per character comparison) for the "Url starts with" features specified in Section~\ref{sec:feat-flow}. 

Given a URL and referrer of average length $S^\text{url}_\text{avg} \approx S^\text{ref}_\text{avg} \approx 150$ (computed from our data), the lower bound for the number of operations in the feature extraction of the full feature set defined in Section~\ref{sec:featsets} is
%
$N = 26 \times S^\text{url}_\text{avg} + 26 \times S^\text{ref}_\text{avg} + 2 \times (S^\text{url}_\text{avg} - 2) + 6 + 25 = 8127$
%
with most expensive operation being the hash-map lookup with linear complexity $O(S)$ for the worst case of hashing, with $S$ being the number of lookup queries. Because the complexity of feature extraction related to all the proxy log fields except URL and referrer is constant, i.e. $O(1)$, the complexity of the feature extraction is linear, i.e. $O(S)$, with $S = S^\text{url} + S^\text{ref}$. 
In our case, the maximum number of operations performed in the detection is $T \times D_\text{max} \times L = 40 \times 19 \times 3 = 2280$. Obviously, in the feature extraction we have to perform at least 3-times more operations than in the detection. Moreover, string-based operations are more expensive than simple binary tests applied in the detection. The zeroth level is used to compensate for this fact, see Section~\ref{sec:workflow}.

In order to get an estimate of absolute running times, we benchmarked the system described in the previous section on a computer with 2.3 GHz Intel Core i7 processor. Specifically, we computed the average time needed to process 100k of flows on the dataset Jan2015. The times needed to load the data into memory were not considered and only single thread was used. Using this setup, the average time to process 100k flows was 27 seconds. To get a better insight into this number, a company with 10k users has approximately 19M of flows per day in average. That is, \textit{the entire day} of the network traffic is processed in $(27 \times 19 \times 10^6/10^5) / 60 = 85.5$ minutes.

\section{Analysis of Features Selected by Random Forests}
\label{sec:anal-feat}
Often, we are interested to know the contribution each feature makes during malware classification. In random forest classifier, the features and decision thresholds form split nodes in each tree, see \eqref{eq:rf-f}.  Therefore, the importance of each feature can be estimated from the number of occurrences of a feature in the nodes of individual trees. That is, we count how many times a specific feature was used for the decision making across all the trees. Top features (sorted in descending order) are listed in Table~\ref{tab:top-feat}. These are reported for the classifiers on the second level of hierarchy, see Fig.~\ref{fig:sys-full}, which are based on the full feature sets, see Table~\ref{tab:DGA-trsets}.

\begin{table}
\centering
\caption{List of top features according to the number of occurrences in the nodes of the random forests for reported second-level classifiers (see Fig.~\ref{fig:sys-full}). Features not related to the URL are blue. See the text for discussion.}
\newcommand{\nonurl}[1]{\textcolor{blue}{\textit{#1}}}
\begin{tabular}{|l|l|}
\hline
\textbf{ENC (C\&C) - L2} & \textbf{PHISH} \\ \hline
Length Of URL Path & URL Path Trigram Histogram (first bin)\\ 
\nonurl{Length Of User Agent} & URL Path Trigram Histogram (last bin) \\
\nonurl{Length Of Mime Type} & \nonurl{HTTP Status} \\ 
\nonurl{Server-Client Bytes}  & \nonurl{Server-Client Bytes}  \\ 
Max. Occurrence Ratio Of Character In URL Path & \nonurl{Length Of User Agent} \\ 
\nonurl{HTTP Status} & Max. Occurrence Ratio Of Character In URL Path \\ 
\nonurl{Flow Duration} & Number Of Occurrences Of '+' In URL Path \\ 
Number of Occurrences Of '/' In URL Path & URL Char. Type Ratio Of Upper Case In URL Path \\ 
URL Path Trigram Histogram (first bin) & URL Char. Type Ratio Of Digits In URL Path \\ 
URL Char. Type Ratio Of Lower Case In URL Path & Number Of Occurrences Of '/' In URL Path \\ 
Max. Length Of Stream Of Lower Case In URL Path  & \nonurl{Length Of Mime Type} \\ 
\nonurl{Client-Server Bytes}  & Length Of URL Path \\  
Max. Length Of Stream Of Digits In URL Path & Number Of Occurrences Of '.' In Hostname \\ 
\nonurl{Number Of Occurrences Of '.' In Referrer} & \nonurl{Client-Server Bytes}  \\ 
Max. Length Of Stream Of Lower Case In URL Path & \nonurl{Flow Duration} \\ 
Number Of Occurrences Of '=' In URL Query & \\
Non Letter Ratio Of URL Path & \\ \hline \hline
\textbf{CLICK-FR}  & \textbf{DGA - L2}\\ \hline
Number Of Occurrences Of '.' In Hostname & \nonurl{Length Of User Agent} \\ 
Number Of Occurrences Of '-' In URL Query & \nonurl{Flow Duration} \\ 
URL Char. Type Ratio Of Digits For URL Query & Non Letter Ratio Of URL Path \\ 
URL Path Trigram Histogram (first bin) & \nonurl{Server-Client Bytes}  \\ 
Length Of URL Path & Length Of URL Query\\ 
Max. Length Of Stream Of Digits For URL Query & \nonurl{Length Of Mime Type} \\ 
Number Of Occurrences Of '-' In URL Path & \nonurl{Client-Server Bytes}  \\  
\nonurl{Length Of User Agent} & \nonurl{HTTP Status} \\ 
URL Char. Type Ratio Of Lower Case In URL Path & SLD Trigram Histogram (last bin) \\ 
Length Of URL Query & Vowel Change Ratio Of SLD \\ 
\nonurl{Flow Duration} & Max. Length Of Stream Of Consonants For SLD \\ 
URL Query Trigram Histogram (last bin) & Max. Length Of Stream Of Digits In URL Path \\ 
\nonurl{HTTP Status} & Max. Occurrence Ratio Of Character For SLD \\ 
\nonurl{Length Of Mime Type} & Length Of SLD \\ 
Non Letter Ratio Of URL Query & Length Of URL Path \\
\hline \hline
\end{tabular}
\label{tab:top-feat}
\end{table}

The first observation we can make relates to the way URL and non-URL features (shown in blue) are used. The first level classifiers, as they were proposed, focus only on the domain and URL to filter out proxy logs without any encrypted strings in their URLs or domain names clearly not generated by a DGA. 
It is therefore interesting to see, that while the URL features are still present at the second classification level, each of the second level classifiers strongly relies also on non-URL features. This is most obvious for the ENC and DGA detector. Importantly, the classifiers heavily rely on the number of transferred bytes along with the duration (\emph{Flow duration}) and the HTTP status of the connection. This means that the features related to other fields in the proxy log tend to be more discriminative than the URL and domain-related properties alone when classifying C\&C communication. This further explains the typical FPs when the system relies solely on URLs as shown in Table~\ref{tab:tunn-ex} and Table~\ref{tab:dga-ex}.

Another set of important features is related to lengths of various proxy log fields (\emph{Length of ...} features), which are used to separate empty or missing fields or short user agent strings (common indicator of compromise\cite{Referrer2011}) important for DGA and ENC detection.
The features \emph{Number of occurrences of '.' in hostname} are used to distinguish hostnames having many subdomains or when IP address is used instead of the domain name (e.g.\ hxxp://123.123.123.123/path/). It is also interesting that only the values related to the first and the last bin of the histogram-based features are in the top of the list. These values represent the count of least and most common trigrams occurring in the input string (URL Path/Query/SLD).

For ENC-L2 detector, the most common features is \emph{Length of URL path} since URLs with a long path and encryption (detected by the first classification level) are not common in legitimate traffic. There are also other features, which discriminate well between parts of the legitimate and malicious traffic such as counting the number of slashes in the URL path (\emph{Number of Occurrences Of '/' In URL Path}) and equal signs in the query (\emph{Number Of Occurrences Of '=' In URL Query}). Their values are then compared with the expected numbers of these characters using the thresholds in the respective nodes.

The top features for the phishing classifier are the counts of the least and the most common trigrams in the URL path. This means that the classifier distinguishes between URLs with rare and common character distributions. However, the classifier makes further fine-grained decision based on three non-URL features: \emph{HTTP Status} (redirections are often used as a part of the phishing scam), \emph{Server-Client Bytes} and \emph{Length Of User Agent}. 

Unlike the phishing classifier which focused narrowly on the URL path, the click-fraud classifier relies on various properties of the URL string (lengths, different character distributions, trigram counts of both URL query and path). In addition, several non-URL features are used out of which the length of the user agent plays the most important role.

For the DGA-L2 classifier, none of the most important features is related to the second level domain and its character distribution. Clearly, the second classification level compensates for the lack of information provided on the first level to further differentiate the malicious and legitimate communication patterns. The classifier uses mainly different parts of the URL (such as path and query) and the other proxy log fields.

Finally, it is important to remember that none of these generic features is employed in pattern matching of specific strings. These features capture statistics, combinations, relationships, and dependencies which would be very difficult to derive manually by a human expert. All these characteristics were discovered automatically from training data by the learning algorithm, which additionally provides scoring of the features according to their relevance. This makes it possible to tailor classifiers to the individual malware families while reusing the common architecture and the learning components.

\subsection{Evasion}
The evasion of our system would involve modifying the proxy logs to alter the important features (as identified in Table~\ref{tab:top-feat}) and cause misclassification of malicious traffic. For example, an attacker might modify the user agent to alter the feature \emph{Length of user agent}. One potential remedy could be to remove the feature from training process in order to increase the robustness of the system, or to enlarge the training set by malicious samples having higher variation in lengths of the user agent.

It is important to realize, however, that altering one feature will not evade the system. The random forest classifier consists of several trees, each having several different paths that lead to the detection of a malicious sample. When the trees are learned, the robustness is increased by taking \textit{random subsets} of the data and features. This means that altered feature might not even be selected for training. During classification, the overall detection decision is then given by the majority vote from all the trees, see Section~\ref{sec:rf}. Therefore, if an attacker alters a specific feature (e.g.\ length of the user agent), although some of these votes would not be captured, the overall decision would be the same in most cases. The detection failures could be used to retrain the system to adapt to the new conditions. Furthermore, the full system can be easily retrained with a different random seed which would produce different trees with different detection rules along their paths. This would render the evasion ineffective and presents a major advantage when compared to manually constructed rules.

\lmanew{The majority of deployed intrusion detection systems still rely on hand-crafted patterns or blacklists. The proposed system extends the existing intrusion detection capabilities by providing an alternative to the expert-defined rules through a powerful learning architecture. The learning system can process hundreds of generic features, find the most relevant correlations of features, and generalize the traffic patterns -- all within a limited scope of a single proxy log.}

\section{Related Work}
\label{sec:rel-work}
%
Network-based Intrusion Detection Systems (NIDS) are deployed on various networks segments, typically at the edge of the network, to monitor all incoming and outgoing network traffic \cite{ModiJNCA2012,LiaoJNCA2012}. The systems use different strategies to detect malicious communication which can be broadly characterized into systems analyzing payload or payload statistics, systems processing proxy logs or Netflows, systems that build rules or signatures in a separate controlled environment, and systems relying on additional data sources such as DNS records.

Payload packet inspection has been proposed in the PAYL system \cite{PAYL}, where the fingerprints are represented as statistical distributions of bytes in a payload. The Provex system \cite{Provex} extends the inspection to encrypted data by maintaining a list of known decryption methods. In BotHunter \cite{BotHunter}, the anomalies found by PAYL are combined with anomaly detections from inbound and outbound scans and correlated with intrusion reports based on Snort rules \cite{Snort}.
BotSniffer \cite{BotSniffer} computes the anomaly scores from peaks in the number of entries for each destination IP and port pair aggregated over a longer time period.
The anomaly detectors are reused in the BotMiner system \cite{BotMiner} to create clusters of similar host activities (so called A-plane). 
The A-plane is correlated with the clustering of communication patterns represented by the flow statistics (C-plane) to produce the final report.
Despite their effectiveness, the packet analysis NIDS systems cannot always be deployed since the increasing network bandwidth increases the rates of data processing and packet inspection may violate privacy protection imposed by internal local network policies.
The proposed system does not use packet inspection or clustering which allows it to achieve fast processing speeds necessary for online deployment.

Other types of IDS are based on deriving signatures directly from malware samples run in controlled environments. In \cite{Perdisci2010}, statistics from all the HTTP traffic are collected and clustered based on simple statistical HTTP features and similarity of URLs. Similar clusters are merged which results in a more compact representations. Signatures in the Snort format are derived from each merged cluster based on substrings that are shared for all the samples in a cluster. This approach is extended in the Firma \cite{Firma} system to use all traffic generated by the malware, not only HTTP traffic.
The signatures in the BotZilla \cite{BotZilla} framework are manually derived from tokenized substrings of several proxy log strings, such as user agent and the URL. Substrings which occur at least $d$ times in a training set and were not seen more than $n$ times in a legitimate traffic are used to define signatures.
Extracting signatures from malware samples run in a controlled environment has limitations. According to the analysis in \cite{Sandnet}, only 44\% of malware samples exhibits a network activity, out of which 59\% is HTTP. Moreover, only 6\% of the flows starts in the first 5 minutes of the malware activity and only 24\% of communication endpoints is contacted during that time. These statistics underscored by the fact that the malware behaves differently in controlled environments restricts the applicability of the generated signatures.
Our approach can be seen as a data-driven alternative to complement signature generation in cases, where signatures are hard or impossible to define. 

Some Intrusion Detection Systems (IDS) rely on passive DNS analysis to detect malicious domains. The Exposure system \cite{Exposure} monitors long time periods, from which distinct DNS statistics are extracted and used to train a C4.5 decision tree. 
DNS request logs have also been used in NIDS systems detecting communication with malicious domains produced by DGA. 
Pleiades \cite{Antonakakis2012} detector analyzes unsuccessful domain name resolutions (NXDOMAIN responses). The responses are collected over one day and clustered and classified. However, the requirement is that the unsuccessful domain name resolutions are available as the input.
Although we do not rely on passive DNS analysis in our system, it could be easily exploited by introducing an anomaly detection layer which would pre-filter flows that are then analyzed by the classification system.

The investigation of the proxy log URLs \cite{McGrathLEET2008} prompted the use of host-based and lexical features to train various statistical detection models \cite{MaKDD2009,ChoiUSE2011,Zhao2013,HuangSVM2012,MaACM2011,KurtISSP2011}.
The lexical information is captured by creating a dictionary of URL parts tokenized according to a predefined set of delimiters. The URL is then represented as a collection of binary feature vectors which specify the indices of the corresponding dictionary tokens (so called bag-of-words model). Since the number of unique tokens is large, the dimension of the resulting feature vector is in the order of millions \cite{MaACM2011,ChoiUSE2011,KurtISSP2011}. Host-based features use IP address, DNS and WHOIS records to further support the decision making. Some systems use also the context-based features related to the HTML content, JavaScript events, page links or iFrames \cite{ProvosUSENIX2008,WhittakerNDSS2010,ChoiUSE2011} which can result in large feature vectors, up to 50 million dimensions \cite{KurtISSP2011}. One disadvantage of these approaches is that the bag-of-words model has to be frequently updated with new tokens \cite{MaICML09}.
Although new tokens can be incrementally added to the existing model \cite{MaICML09,MaACM2011}, reliable malware feeds have to be available to indicate that the new tokens are related to malware, or specific malware type \cite{ChoiUSE2011}.
The labeling cost can be reduced by querying only a small portion of the detected samples such that the labeling budget is met \cite{Zhao2013}.
Our system uses generic features extracted from the proxy log URLs along with the other log attributes. Together, they form a compact representation that reflects the differences between malicious and legitimate behaviors. We do not use features related to the content of a webpage the URL is pointing to. The main arguments are that downloading information content of a webpage with related sources, both for training and testing, can be from practical point of view troublesome and it can further slow down the classification process \cite{MaKDD2009}.

The domain names of the proxy log URLs can also be used to compute statistical features for detecting malware communication with the DGA domains. These features include n-gram frequency distributions, character counts and count ratios, character entropy, presence in a dictionary and other measures. The features are often combined with additional indicators such as various rules on the proxy log fields, or feeds blacklisting/whitelisting domain names. All the evidence is then evaluated automatically with a rule-based \cite{Scarfone2007,Gregoire2008} or machine learning-based algorithms \cite{DomFlux2012,Antonakakis2012,Phoenix}. 
In \cite{DomFlux2012}, the detection precision is increased by clustering the domain names according to their corresponding serverIPs, or by discarding small clusters of unique sub-domains found for each second level domain. Unfortunately, the recall of the system can decrease by incorrectly discarding clusters with malicious communication. The final detection is based on the similarity of domains in each group.
The Phoenix \cite{Phoenix} system uses a pre-filtering step in which domains with non-standard character distribution are filtered and subsequently clustered according to the IP addresses they resolved to. Pruning and merging of clusters is performed to decrease the number of resulting clusters and increase the precision.
The proposed system detects proxy logs with DGA domains based on statistical features extracted from the domain characters. Additional features from the proxy logs improve the detector precision since more evidence is used in the decision.


Our detection system classifies individual flows and therefore can report the threats immediately as they occur. Compared to clustering approaches, this has the advantage that the system does not need several potential infections that are otherwise needed to initiate the clustering. In addition, the computational costs are much lower \cite{Bitshred,Perdisci2010}. 

Our system detects malicious communication patterns represented by generic features. The features are computed from all proxy log fields (except client/server IP) and a supervised classifier is trained. Therefore, the system does not require any knowledge gained from reverse engineering of a botnet. 

Our architecture can easily incorporate new threats by retraining the classifiers.
As shown in Section~\ref{sec:phish-exp}, the retraining can use a small set of samples which is not possible with clustering approaches that remove them as not interesting \cite{BotSniffer,BotMiner,Phoenix,Perdisci2010}.
The multi-level design makes it possible to retrain individual levels by identifying false negatives from the analysis of the inputs and outputs at various levels.

\section{Conclusions}
The proposed discriminative learning system trains classifiers of proxy log records to detect malicious network traffic. Each classifier in the multi-stage hierarchy can be retrained independently whenever new samples are available. The key advantage that decisions are performed independently for each proxy log without using any additional information may limit the scope of the detection system. Our ongoing work focuses on the architecture extension by incorporating the traffic context. This can be done by training new base classifiers that operate within a given time window or on a sequence of flows to a domain, host, or an IP address. These classifiers can then be used at various levels of the multi-level architecture.

\bibliographystyle{elsarticle-num}
\bibliography{mybib-rf}

\end{document}